\documentclass[10ptjournal]{IEEEtran}
\usepackage{mathtools}
\usepackage{amsmath}
\usepackage{cleveref}
\ifCLASSINFOpdf
  \usepackage[pdftex]{graphicx}
  \graphicspath{{../pdf/}{../jpeg/}}
  \DeclareGraphicsExtensions{.pdf,.jpeg,.png}
\else
   \usepackage[dvips]{graphicx}
\fi
\usepackage{bbm}

\DeclarePairedDelimiter\norm{\lVert}{\rVert}
\makeatletter
\let\oldabs\abs
\def\abs{\@ifstar{\oldabs}{\oldabs*}}
\let\oldnorm\norm
\def\norm{\@ifstar{\oldnorm}{\oldnorm*}}
\makeatother

\usepackage{calc}
\usepackage{amssymb}
\usepackage{easybmat}
\usepackage{algpseudocode}
\usepackage{algorithm}
\usepackage{amsmath} % assumes amsmath package installed
\usepackage{amsthm}

\begin{document}
%
% paper title
% can use linebreaks \\ within to get better formatting as desired
% Do not put math or special symbols in the title.
\title{One-Class Kernel Spectral Regression}
%
%
% author names and IEEE memberships
% note positions of commas and nonbreaking spaces ( ~ ) LaTeX will not break
% a structure at a ~ so this keeps an author's name from being broken across
% two lines.
% use \thanks{} to gain access to the first footnote area
% a separate \thanks must be used for each paragraph as LaTeX2e's \thanks
% was not built to handle multiple paragraphs
%

\author{Shervin~Rahimzadeh~Arashloo\IEEEmembership{} and Josef~Kittler        % <-this % stops a space
\IEEEcompsocitemizethanks{\IEEEcompsocthanksitem \protect\\
% note need leading \protect in front of \\ to get a newline within \thanks as
% \\ is fragile and will error, could use \hfil\break instead.
E-mail: 
}% <-this % stops an unwanted space
}
\author{Shervin~Rahimzadeh~Arashloo\IEEEmembership{}
        and~Josef~Kittler,~\IEEEmembership{Life Member}
        % <-this % stops a space
\thanks{S.R. Arashloo is with the department
of computer Engineering, Bilkent university, Ankara, Turkey, 06800. E-mail: s.rahimzadeh@cs.bilkent.edu.tr}% <-this % stops a space
\thanks{J. Kittler is with CVSSP, university of Surrey, Guildford, Surrey, UK, GU2 7XH. E-mail: j.kittler@surrey.ac.uk.}% <-this % stops a space
}

% note the % following the last \IEEEmembership and also \thanks - 
% these prevent an unwanted space from occurring between the last author name
% and the end of the author line. i.e., if you had this:
% 
% \author{....lastname \thanks{...} \thanks{...} }
%                     ^------------^------------^----Do not want these spaces!
%
% a space would be appended to the last name and could cause every name on that
% line to be shifted left slightly. This is one of those "LaTeX things". For
% instance, "\textbf{A} \textbf{B}" will typeset as "A B" not "AB". To get
% "AB" then you have to do: "\textbf{A}\textbf{B}"
% \thanks is no different in this regard, so shield the last } of each \thanks
% that ends a line with a % and do not let a space in before the next \thanks.
% Spaces after \IEEEmembership other than the last one are OK (and needed) as
% you are supposed to have spaces between the names. For what it is worth,
% this is a minor point as most people would not even notice if the said evil
% space somehow managed to creep in.

% The paper headers
\markboth{}%
{Shell \MakeLowercase{\textit{et al.}}: Bare Demo of IEEEtran.cls for Journals}
% The only time the second header will appear is for the odd numbered pages
% after the title page when using the twoside option.
% 
% *** Note that you probably will NOT want to include the author's ***
% *** name in the headers of peer review papers.                   ***
% You can use \ifCLASSOPTIONpeerreview for conditional compilation here if
% you desire.

% If you want to put a publisher's ID mark on the page you can do it like
% this:
%\IEEEpubid{0000--0000/00\$00.00~\copyright~2012 IEEE}
% Remember, if you use this you must call \IEEEpubidadjcol in the second
% column for its text to clear the IEEEpubid mark.

% use for special paper notices
%\IEEEspecialpapernotice{(Invited Paper)}

% make the title area
\maketitle

% As a general rule, do not put math, special symbols or citations
% in the abstract or keywords.
\begin{abstract}
The paper introduces a new efficient nonlinear one-class classifier formulated as the Rayleigh quotient criterion optimisation. The method, operating in a reproducing kernel Hilbert space, minimises the scatter of target distribution along an optimal projection direction while at the same time keeping projections of positive observations distant from the mean of the negative class. We provide a graph embedding view of the problem which can then be solved efficiently using the spectral regression approach. In this sense, unlike previous similar methods which often require costly eigen-computations of dense matrices, the proposed approach casts the problem under consideration into a regression framework which is computationally more efficient. In particular, it is shown that the dominant complexity of the proposed method is the complexity of computing the kernel matrix. Additional appealing characteristics of the proposed one-class classifier are: 1-the ability to be trained in an incremental fashion (allowing for application in streaming data scenarios while also reducing the computational complexity in a non-streaming operation mode); 2-being unsupervised, but providing the option for refining the solution using negative training examples, when available; And last but not the least, 3-the use of the kernel trick which facilitates a nonlinear mapping of the data into a high-dimensional feature space to seek better solutions. 

Extensive experiments conducted on several datasets verify the merits of the proposed approach in comparison with  other alternatives.
\end{abstract}

% Note that keywords are not normally used for peerreview papers.

% For peer review papers, you can put extra information on the cover
% page as needed:
% \ifCLASSOPTIONpeerreview
% \begin{center} \bfseries EDICS Category: 3-BBND \end{center}
% \fi
%
% For peerreview papers, this IEEEtran command inserts a page break and
% creates the second title. It will be ignored for other modes.
\IEEEpeerreviewmaketitle

\section{Introduction}
\IEEEPARstart{O}{ne}-class classification (OCC) \cite{khan_madden_2014} deals with the problem of identifying objects, events or observations which conform to a specific behaviour or condition, identified as the target/positive class ($\mathcal{T}$), and distinguishing them from all other objects, typically known as outliers or anomalies. More specifically, consider a set of points $X = \{x_1, \dots, x_{n} \}$ where $x_i\in\mathbb{R}^d$ is a realisation of a multivariate random variable $x$ drawn from a target probability distribution with probability density function $p(x)$. In a one-class classification problem, the goal is to characterise the support domain of $p(x)$ via a one-class classifier $h(z)$ as
\begin{equation}
 h(z) = \lceil q(z)\leq\tau\rceil=\left\{
  \begin{array}{ll}
    1 &\text{$z\in \mathcal{T}$}\\
    0 &\text{otherwise}
  \end{array} \right.
\end{equation}
\noindent where function $q(.)$ is modelling the similarity of an observation to the target data and $\lceil . \rceil$ denotes the Iverson brackets, returning an output of 1 when the argument is correct and zero otherwise. Parameter $\tau$ is optimised so that an expected fraction of observations lie within the support domain of the target distribution. One-class learning serves as the core of a wide variety of applications such as intrusion detection \cite{6846360}, novelty detection \cite{BEGHI20141953}, fault detection in safety-critical systems \cite{4694106}, fraud detection \cite{Kamaruddin:2016:CCF:2980258.2980319}, insurance \cite{7435726}, health care \cite{6566012}, surveillance \cite{4668357}, network anomaly detection \cite{8386786}, etc. Historically, the first single-class classification problem seems to date back to the work in \cite{Minter1975} in the context of learning Bayes classifier. Later, with a large time gap, the term one-class classification was used in \cite{1993STIN9324043M}. As a result of a widening spectrum of applications of one-class classification, other terminology has been adopted, including anomaly/outlier detection \cite{RITTER1997525}, novelty detection \cite{318023}, concept learning \cite{Japkowicz:1999:CLA:929980}.
\begin{figure}[t]
\centering
\includegraphics[scale=.2]{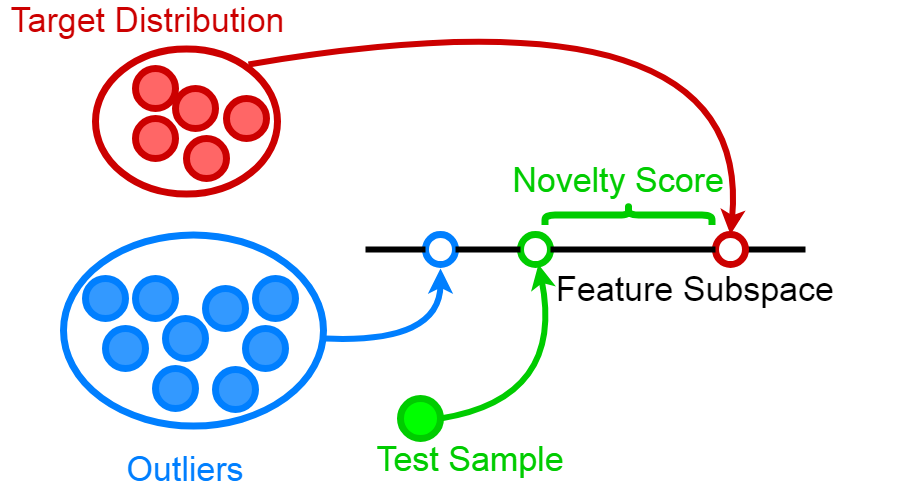}
\caption{Overview of the proposed approach: positive training samples of target distribution and outliers are mapped onto distinct points in an optimal feature subspace. If no outlier training samples exist, the origin would serve as a single artificial outlier. The novelty score of a test sample is defined in terms of the distance between its projection and the projection of target distribution in an optimal feature subspace.}
\end{figure}
OCC techniques are commonly employed when the non-target/negative class is either not well defined, poorly sampled or totally missing, which may be due to the openness of the problem \cite{6809169} or due to the high cost associated with obtaining negative samples. In these situations, the conventional two-class classifiers are believed not to operate as effectively as expected since they are based on the assumption that data from all classes are more or less equally balanced. OCC techniques are developed to address this shortcoming of the conventional approaches by primarily training on the data associated with a \textit{single} class. Nevertheless, the lack of negative samples may pose serious challenges in learning one-class classifiers as the decision boundary should be estimated using only positive observations. As a result, the one-class problem is typically considered to be more difficult than the two-class counterpart. As observed in \cite{30e27ea86}, the challenges related to the standard two/multi-class problems, e.g. estimation of the error, atypical training data, the complexity of a solution, the generalisation capability, etc. are also present in OCC and may sometimes become even more severe.

Although there may exist a fine grain categorisation of one-class techniques \cite{khan_madden_2014,30e27ea86,PIMENTEL2014215}, a general overarching classification considers them to be either generative or non-generative \cite{6636290}. The generative approaches incorporate a model for generating all observations, whereas non-generative methods lack a transparent link to the data. In this context, the non-generative methods are best represented by discriminative approaches which partition the feature space in order to classify an object. As notable representatives of the generative approaches one may consider the parametric and nonparametric density estimation methods \cite{Tax:2001:COC:648055.744087,10.1137/1.9781611973440.67,HOFFMANN2007863} (using for example a Gaussian, a mixture of Gaussians or a Poisson distribution), neural-network based methods \cite{Japkowicz:1999:CLA:929980,7492368}, one-class sparse representation classification \cite{7984788,7393467}, etc. Well-known examples of the non-generative methods include those based on support vector machines (SVDD/one-class SVM) \cite{Tax2004,Scholkopf:2001:ESH:1119748.1119749,6869001}, linear programming \cite{efddef0571dc4c9594328f21683c3d45}, convex hull methods \cite{10.1007/978-3-642-21557-5_13,8125573}, cluster approaches \cite{10.1007/978-1-4471-1599-1_110}, deep-learning based methods \cite{DBLP:journals/corr/abs-1802-09088,DBLP:journals/corr/abs-1801-05365}, extreme learning-based methods \cite{7938706}, ensemble approaches \cite{8293843} and subspace methods \cite{NIPS2004_2656,Roth:2006:KFD:1117520.1117529,6619277,6857384,6384805}. By virtue of the emphasis on classification, rather than modelling the generative process, the non-generative approaches tend to yield better performance in classification \cite{6636290}.

In practical applications where the data to be characterised is highly nonlinear and complex, \textit{linear} approaches often fail to provide satisfactory performance. In such cases, an effective strategy is to implicitly map the data into a very high dimensional space so that in this new space the data become more easily separable, the prominent examples of which are offered by kernel machines \cite{kernelfisher,Gkernel,kernelbook,cortes1995support}. Nevertheless, the high computational cost associated with these methods can be considered as a bottleneck in their usage. For instance, the one-class variants of kernel discriminant analysis \cite{6619277,7727608,6857384,8099922} often require computationally intensive eigen-decompositions of dense matrices.

In this work, a new nonlinear one-class classifier, formulated as optimisation of a Rayleigh quotient, is presented which unlike previous discriminative methods \cite{NIPS2004_2656,Roth:2006:KFD:1117520.1117529,6619277,6857384,6384805,8099922} avoids costly eigen-analysis computations via the spectral regression (SR) technique. This solution has been shown to speed up the kernel discriminant analysis by several orders of magnitude \cite{SRKDA}. By virtue of bypassing the eigen-decomposition of large matrices via a regression formulation, the proposed One-Class Kernel Spectral-Regression (OC-KSR) approach becomes computationally very attractive, with the dominant complexity of the algorithm being relegated to the computation of the kernel matrix. An additional appealing characteristic of the method is the ameanability to be applied in an incremental fashion, allowing for the injection of additional training data into the system in a streaming data scenario, side-stepping the need to reinitialise the training procedure, while also reducing the computational complexity in a non-streaming operation mode. Additionally, the method can be operated in an unsupervised mode as well as by using some negative examples in the training set to further refine the solution.

\subsection{Overview of the Proposed Approach}
In the proposed one-class method, the strategy is to map the data into the feature space corresponding to a kernel such that: 1-the scatter of the data along the projection direction is minimised; 2-the projected samples and the mean of negative class along the projection direction are maximally distant. The problem is then posed as one of graph embedding which is optimised efficiently using the spectral regression technique \cite{SRKDA}, thus avoiding costly eigen-analysis computations. In addition, an incremental version of the proposed method is also presented which reduces the computational complexity of the training phase even further. Although in an OCC problem negative training examples are not always expected to exist, if they do, the proposed method is able to utilise them to further refine the decision boundary. During the test phase, the decision criterion for the proposed approach involves projecting a test sample onto the inferred optimal feature space followed by computing the distance between its projection and that of the mean of the target samples.

The main contributions of the present work are thus summarised as
\begin{itemize}
\item A method of designing a nonlinear one-class classifier (OC-KSR) developed from a graph embedding formulation of the problem;
\item Efficient optimisation of the proposed formulation based on spectral regression;
\item An incremental variant of the OC-KSR approach;
\item An extension of the proposed OC-KSR method to benefit from possible negative samples in the training set in a supervised operating mode;
\item And, an extensive evaluation of the proposed method and its comparison to the state-of-the-art one-class classification techniques on several datasets.
\end{itemize}

\subsection{Outline of the Paper}
The rest of the paper is organised as follows: In Section \ref{related}, the one-class methods which are closely related to the proposed method are reviewed. In doing so, the focus is more on nonlinear methods posing the one-class classification problem as an optimisation of (generalised) Rayleigh quotient. In Section \ref{OC-KSR}, the proposed one-class method (OC-KSR) is presented. An experimental evaluation of the proposed approach along with a comparison to other methods on several datasets is provided in Section \ref{exps}. Finally, the paper is drawn to conclusions in Section \ref{conclude}. 

\section{Related Work}
\label{related}
As an example of the unsupervised methods using a Rayleigh quotient, the work in \cite{HOFFMANN2007863} employs kernel PCA for novelty detection where a principal component in a feature space captures the distribution of the data and the reconstruction residual of a test sample with respect to the inferred subspace is employed as a novelty measure. Other work in \cite{GueSchVis07} describes a strategy to improve the convergence behaviour of the kernel algorithm for the iterative kernel PCA. A different study \cite{4522554} proposed a robustified PCA to deal with outliers in the training set.

In \cite{NIPS2004_2656,6795824}, a one-class kernel Fisher discriminant classifier is proposed which is related to Gaussian density estimation in the induced feature space. The proposed method is based on the idea of separating the data from their negatively replicated counterpart and involved an eigenvalue decomposition of the kernel matrix. In this approach, the data are first mapped onto some feature space where a Gaussian model is fitted. Mahalanobis distance to the mean of this Gaussian is used as a test statistic to test whether the data is explained by the model. As pointed out in \cite{6795824}, for kernel maps which transform the input data into a higher-dimensional space, the assumption that the target data is normally distributed may not hold in general. If the deviation from normality is large, the methods in \cite{NIPS2004_2656,6795824} may lead to unreliable results.

The work in \cite{6619277} proposed a Fisher-based null space method where a zero within-class scatter and a positive between-class scatter are used to map all training samples of one class onto a single point. The proposed method treats multiple known classes jointly and detects novelty with respect to the set of classes using a projection onto a joint subspace where the training samples of all known classes are presumed to have zero variance. Checking for novelty involves computing a distance in the estimated subspace. The method requires eigen-decomposition of the kernel matrix. In a follow-up work \cite{7045967}, it is proposed to incorporate locality in the null space approach of \cite{6619277} by considering only the most similar patterns to the query sample, leading to improvements in performance. In \cite{8099922}, an incremental version of the method in \cite{6619277} is proposed to improve on computational efficiency.

In \cite{6857384,DUFRENOIS201696}, a generalised Rayleigh quotient specifically designed for outlier detection is proposed. The method tries to find an optimal hyperplane which is closest to the target data and farthest from the outliers which requires building two scatter matrices: an outlier scatter matrix corresponding to the outliers and a target scatter matrix for the target data. While in \cite{6857384}, the decision boundary is found by a computationally intensive generalised eigenvalue problem which limits the use of the method to medium sized datasets, in \cite{DUFRENOIS201696}, the generalised eigenvalue problem is replaced by an approximate conjugate gradient solution to decrease the computational cost. The method presented in \cite{6857384,DUFRENOIS201696} has certain shortcomings as the computation of the outlier scatter matrix requires the presence of atypical instances which is sometimes difficult to collect in some real applications. Another drawback is that the method is based on the assumption that the target population differs from the outlier population in terms of their respective densities which might not hold for real-world problems in general. A later study \cite{7727608} tries to address these shortcomings via a null-space variant of the method in \cite{6857384,DUFRENOIS201696}. In order to overcome the limitation of the availability of outlier samples, it is proposed to separate the target class from the origin of the kernel feature space, which serves as an artificial outlier sample. The density constraint is then relaxed by deriving a joint subspace where the training target data population have zero covariance. The method involves eigen-computations of dense matrices.

While the majority of previous work on one-class classification using a Rayleigh quotient formulation requires computationally intensive eigen-decomposition of large matrices, in this work, a one-class approach is proposed which replaces costly eigen-analysis computations by the spectral-regression technique \cite{SRKDA}. In this sense, the present work can be considered as a one-class variant of the multi-class approach in \cite{SRKDA} and the two-class, class-specific method of \cite{6905848} with additional contributions discussed in the subsequent sections.

\section{One-Class Kernel Spectral Regression}
\label{OC-KSR}
\begin{table}
\footnotesize
\renewcommand{\arraystretch}{1.2}
\caption{Summary of Notations}
\label{notations}
\centering
\begin{tabular}{ c l }
\hline
\textbf{Notation}& \textbf{Description}\\
\hline
$\mathcal{T}$ & The target class\\
$n$ & Total number of training samples\\
$n_0$ & Number of labelled negative examples in the training set\\
$x_i$ & The $i^{th}$ observation in the training set\\
$d$ & Dimensionality of observations in the input space\\
$\mathcal{F}$& The feature (reproducing kernel Hilbert) space\\
$\boldsymbol\phi(.)$ & The nonlinear mapping function onto the feature space\\
$S(\mathcal{T})$ & Scatter of positive training observations along projection direction\\
$\mathcal{M}$ & The mean of projected positive samples\\
$f(.)$ & The projection function\\
$\mathbb{R}$& The set of real numbers \\
$\mathbb{R}^d$ & The set of real vectors in the $d$-dimensional space \\
$\mathbf{E}$& Graph adjacency matrix\\
$\mathbf{I}$& The identity matrix\\
$\mathbf{1}$& A matrix of 1's\\
$\mathbf{L}$ & Graph Laplacian matrix\\
$\mathbf{D}$ & Graph degree matrix\\
$B(\mathcal{T})$ & Sum of squared distances of positive training observations to the\\
								&mean of the non-target class\\
$\boldsymbol\alpha$ & The transformation vector \\
$\mathbf{y}$ & The vector of responses (projections)\\
$S_b$ & Between-class scatter\\
$S_w$ & Within-class scatter\\
$\mathbf{K}$ & The kernel matrix \\
$\kappa(.,.)$ & The kernel function\\
$\tau$ & The threshold for deciding normality \\
$\delta$ & The regularisation parameter\\
\hline
\end{tabular}
\normalsize
\end{table}

Let us assume that there exist $n$ samples $x_1,x_2,\dots x_n \in \mathbb{R}^d$ and $\mathcal{F}$ is a feature space (also known as RKHS:reproducing kernel Hilbert space) induced by a nonlinear mapping $\boldsymbol\phi: \mathbb{R}^d \rightarrow \mathcal{F}$. For a properly chosen mapping, an inner product $\langle.,.\rangle$ on $\mathcal{F}$ may be represented as $\langle \boldsymbol\phi(x_i),\boldsymbol\phi(x_j)\rangle = \kappa(x_i,x_j)$, where $\kappa(.,.)$ is a positive semi-definite kernel function. Our strategy for outlier detection is to infer a nonlinear mapping onto the feature space induced by $\boldsymbol\phi(.)$ based on two criteria: 1-minimising the scatter of mapped target data in the RKHS along the projection direction; and, 2-maximising their distances from the mean of non-target observations in this space. In doing so, the problem is formulated as one of graph embedding which is then posed as optimising a Rayleigh quotient. The optimisation problem is then efficiently solved using spectral regression. The two criteria used in this work to find an optimal subspace are discussed next.

\subsection{Scatter in the feature subspace}
Let us consider a projection function $f(.)$ which maps each target data point $x_i$ onto the feature space. For the reasons to be clarified later, $f(.)$ is assumed to be a one-dimensional mapping. The scatter of target data ($\mathcal{T}$) in the feature space along the direction specified by $f(.)$ is defined as
\begin{equation}
S(\mathcal{T})=\sum_{i=1}^n (f(x_i)-\mathcal{M})^2
\label{var}
\end{equation}
\noindent where $\mathcal{M}$ denotes the mean of all projections $f(x_i)$'s, i.e.
\begin{equation}
\mathcal{M}=\frac{1}{n}\sum_{i=1}^n f(x_i)
\label{M}
\end{equation}
In order to detect outliers, it is desirable to find a projection function $f(.)$ which minimises dispersion of positive samples and forms a compact cluster, i.e minimises $S(\mathcal{T})$. $f(.)$ can be written in terms of real numbers $\alpha_{i}$'s and a positive definite kernel function $\kappa(.,.)$ defining an $n \times n$ kernel matrix $\mathbf{K}$ (where $K_{ij}=\kappa(x_i,x_j)$) according to the Representer theorem  \cite{10.1007/3-540-44581} as
\begin{equation}
f(z)\in\{\sum_{i=1}^{n}\alpha_i \kappa(z,x_i)\lvert \alpha_i\in\mathbb{R}\}
\end{equation}

Assuming that the kernel function $\kappa(.,.)$ is chosen and fixed, the problem of minimising $S(\mathcal{T})$ with respect to $f(.)$ gets reduced to 
finding $\boldsymbol{\alpha}^{opt}$, i.e.

\begin{IEEEeqnarray}{l}
\nonumber \min S(\mathcal{T})=\min_{f(.)}\sum_{i=1}^n (f(x_i)-\mathcal{M})^2\\
\nonumber =\min_{\boldsymbol\alpha}\sum_{i=1}^{n}(\sum_{j=1}^n\alpha_j \kappa(x_i,x_j)- \mathcal{M})^2\\
\end{IEEEeqnarray}
\subsubsection{Graph Embedding View}
\begin{figure}[t]
\centering
\includegraphics[scale=.37]{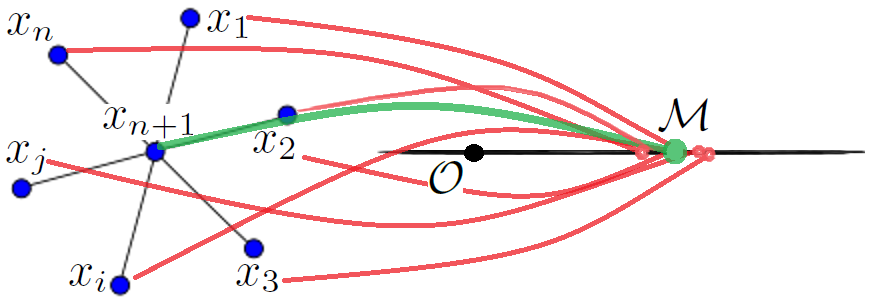}
\caption{The star graph representation of the problem. Left: data points in the original $\mathbb{R}^d$ space; Right: Embedding of the graph onto a line in the feature space. Note: in an optimal projection, all $x_i$'s would be mapped onto $\mathcal{M}$.}
\label{stargraph}
\end{figure}

Let us now augment the dataset ($x_i$'s) with an additional point $x_{n+1}$ satisfying $f(x_{n+1})=\mathcal{M}$. Let us also define the $(n+1)\times(n+1)$ matrix $\mathbf{E}$ as
\begin{eqnarray}
 \mathbf{E}=\left( \begin{matrix} 
	0  & \dots & 0 & 1\\
	\vdots & \ddots & \vdots  & \vdots\\
		 0& \dots & 0& 1\\
	 1& \dots & 1& 0
   \end{matrix}\right)
	\label{matE}
\end{eqnarray}
The scatter $S(\mathcal{T})$ in Eq. \ref{var} can now be written as
\begin{equation}
S(\mathcal{T})=\frac{1}{2}\sum_{i=1}^{n+1} \sum_{j=1}^{n+1}(f(x_i)-f(x_j))^2E_{ij}
\label{var2}
\end{equation}
\noindent where $E_{ij}$ denotes the element of $\mathbf{E}$ in the $i^{th}$ row and $j^{th}$ column. The latter formulation corresponds to a graph embedding view of the problem where the data points are represented as vertices of a graph and $\mathbf{E}$ is the graph adjacency matrix, encoding the structure of the graph. That is, if $E_{ij}=1$, then the two vertices $i$ and $j$ of the graph are connected by an edge. With this perspective and $\mathbf{E}$ given by Eq. \ref{matE}, each data point $x_i, \mbox{for } i=1,\dots,n$ is connected by an edge to $x_{n+1}$, resulting in a \textit{star} graph structure, Fig. \ref{stargraph}. The purpose of graph embedding is to map each node of the graph onto a subspace in a way that the similarity between each pair of nodes is preserved. In view of Eq. \ref{var2}, the objective function encodes a higher penalty if two connected vertices are mapped to distant locations via $f(.)$. Consequently, by minimising $S(\mathcal{T})$, if two nodes are neighbours in the graph (i.e. connected by an edge), then their projections in the new subspace are encouraged to be located in nearby positions.
Defining the diagonal matrix $\mathbf{D}$ such that $D_{ii}=\sum_{j=1}^{n+1}E_{ij}$ would yield
\begin{eqnarray}
 \mathbf{D}=\left( \begin{matrix} 
	1  & 0 & \dots  & 0\\
	0 & \ddots & \vdots  & \vdots\\
		\vdots & \dots & 1& 0\\
	 0& \dots & 0& n
   \end{matrix}\right)
	\label{matD}
\end{eqnarray}
Assuming $\mathbf{y}=(f(x_1), \dots, f(x_{n+1}))$, Eq. \ref{var2} can now be written in a matrix form as
\begin{eqnarray}
\nonumber S(\mathcal{T})&=&\frac{1}{2}\sum_{i=1}^{n+1} \sum_{j=1}^{n+1}(f(x_i)-f(x_j))^2E_{ij}\\
&=& \nonumber \sum_{i=1}^{n+1}f(x_i)D_{ii}f(x_i)-\sum_{i=1}^{n+1}\sum_{j=1}^{n+1}f(x_i)E_{ij}f(x_j)\\
&=& \mathbf{y}^{\top}\mathbf{D}\mathbf{y}-\mathbf{y}^{\top}\mathbf{E}\mathbf{y}
\label{var3}
\end{eqnarray}
\noindent Defining matrix $\mathbf{L}$ as $\mathbf{L}=\mathbf{D}-\mathbf{E}$, Eq. \ref{var3} becomes
\begin{eqnarray}
S(\mathcal{T})=\mathbf{y}^{\top}\mathbf{L}\mathbf{y}
\label{s4}
\end{eqnarray}
\noindent In the graph embedding literature, $\mathbf{D}$ is called degree matrix, the diagonal elements of which counts the number of times an edge terminates at each vertex while $\mathbf{L}$ is graph Laplacian \cite{8294302,8047276}. Since our data points are connected to an auxiliary point $x_{n+1}$ in the star graph representation, minimising the scatter given by Eq. \ref{s4} with respect to projections of target observations (i.e. with respect to $y_i$ for $i=1,\dots,n$) forces the projections to be located in nearby positions to $f(x_{n+1})$. As $f(x_{n+1})=\mathcal{M}$ is the mean of data in the RKHS, by minimising $S(\mathcal{T})$ all target data are encouraged to be as close as possible to their mean on a line defined by $f(.)$ in the feature space. The optimum of the objective function $S(\mathcal{T})$ would be reached if all target data are exactly mapped onto a single point, i.e. $\mathcal{M}$.

\subsection{Origin as an artificial outlier}
The idea of using the origin as an exemplar outlier has been previously used in designing one-class classifiers such as OC-SVM \cite{Scholkopf:2001:ESH:1119748.1119749} and others \cite{7727608,6619277,8099922}. In a sense, such a strategy corresponds to the assumption that novel samples lie around the origin while target objects are farther away. In \cite{Scholkopf:2001:ESH:1119748.1119749}, it is shown that using a Gaussian kernel function, the data are always separable from the origin. In this work, a similar assumption is made and target data points are mapped onto locations in a feature subspace such that they are distant from the origin. In order to encourage the mapped data points to lie at locations far from the origin in the subspace, we make use of sum of squared (Euclidean) distances between the projected data points and the origin. As the projection of the origin in the feature space onto any single subspace (including the one specified by $f(.)$) would be zero, the sum of squared distances of projected data points to the projection of the origin on a subspace defined by $f(.)$ can be written as
\begin{equation}
B(\mathcal{T})=\sum_{i=1}^{n} f(x_i)^2
\label{dis1}
\end{equation}
\noindent and using a vector notation, one obtains
\begin{equation}
B(\mathcal{T})=\mathbf{y}_{-}^{\top}\mathbf{y}_{-}
\label{dis2}
\end{equation}
\noindent where $\mathbf{y}_{-}=(f(x_1),\dots,f(x_n))$ is obtained by dropping the last element of $\mathbf{y}$ which corresponds to our augmented point. As per definition of $B(\mathcal{T})$, its maximisation corresponds to maximising the average margin between the projected target data points and the exemplar outlier.
\subsection{Optimisation}
We now combine the two criteria corresponding to minimising the scatter while maximising the average margin and optimise it with respect to the projections of all target data, i.e. with respect to $\mathbf{y}_{-}=(f(x_1),\dots,f(x_n))$, as
\begin{eqnarray}
\nonumber \mathbf{y}_{-}^{opt} &=& \operatorname*{arg\,min}_{\mathbf{y}_{-}}\frac{S(\mathcal{T})}{B(\mathcal{T})}=\operatorname*{arg\,min}_{\mathbf{y}_{-}}\frac{\mathbf{y}^{\top}\mathbf{L}\mathbf{y}}{\mathbf{y}_{-}^{\top}\mathbf{y}_{-}}\\
&=&\operatorname*{arg\,min}_{\mathbf{y_{-}}}\frac{\mathbf{y}^{\top}\mathbf{D}\mathbf{y}-\mathbf{y}^{\top}\mathbf{E}\mathbf{y}}{\mathbf{y}_{-}^{\top}\mathbf{y}_{-}}
\label{ray}
\end{eqnarray}
Note that the numerator of the quotient is defined in terms of $\mathbf{y}$ whereas the optimisation is performed with respect to $\mathbf{y}_{-}$. Thus, the numerator needs to be expressed in $\mathbf{y}_{-}$. Regarding $\mathbf{y}^{\top}\mathbf{E}\mathbf{y}$ we have
\begin{eqnarray}
\nonumber \mathbf{y}^{\top}\mathbf{E}\mathbf{y}&=&(y_1,\dots,y_{n+1})\left( \begin{matrix} 
	0  & \dots & 0 & 1\\
	\vdots & \ddots & \vdots  & \vdots\\
		 0& \dots & 0& 1\\
	 1& \dots & 1& 0
   \end{matrix}\right)(y_1,\dots,y_{n+1})^{\top}\\
	\nonumber &=&(y_1,\dots,y_{n+1}) (y_{n+1},\dots,y_{n+1},\sum_{i=1}^n y_i)^{\top}\\
	\nonumber &=&y_{n+1}(\sum_{i=1}^n y_i)+y_{n+1}(\sum_{i=1}^n y_i)\\
	\nonumber &=&\frac{2}{n}(\sum_{i=1}^n y_i)(\sum_{i=1}^n y_i)\\
	\nonumber &=& \frac{2}{n}(\mathbf{y}_{-}^{\top}\mathbf{1}^{n\times 1})(\mathbf{1}^{1\times n}\mathbf{y}_{-})\\
	          &=& \frac{2}{n}\mathbf{y}_{-}^{\top}\mathbf{1}^{n\times n}\mathbf{y}_{-}
\end{eqnarray}
\noindent where $\mathbf{1}^{n\times n}$ denotes an $n\times n$ matrix of 1's.

Due to the special structure of $\mathbf{D}$ given in Eq. \ref{matD}, for $\mathbf{y}^{\top}\mathbf{D}\mathbf{y}$, one obtains
\begin{eqnarray}
\nonumber \mathbf{y}^{\top}\mathbf{D}\mathbf{y}&=&\mathbf{y}_{-}^{\top}\mathbf{y}_{-}+n(f(x_{n+1}))^2=\mathbf{y}_{-}^{\top}\mathbf{y}_{-}+n(\frac{1}{n}\sum_{i=1}^n y_i)^2\\
\nonumber &=&\mathbf{y}_{-}^{\top}\mathbf{y}_{-}+\frac{1}{n}(\sum_{i=1}^n y_i)^2\\
\nonumber &=& \mathbf{y}_{-}^{\top}\mathbf{y}_{-}+\frac{1}{n}(\mathbf{y}_{-}^{\top}\mathbf{1}^{n\times 1})(\mathbf{1}^{1\times n}\mathbf{y}_{-})\\
&=& \mathbf{y}_{-}^{\top}\mathbf{y}_{-}+\frac{1}{n}\mathbf{y}_{-}^{\top}\mathbf{1}^{n\times n}\mathbf{y}_{-}
\end{eqnarray}
As a result, Eq. \ref{ray} can be purely written in terms of $\mathbf{y}_{-}$ as
\begin{eqnarray}
\nonumber \mathbf{y}_{-}^{opt} &=& \operatorname*{arg\,min}_{\mathbf{y}_{-}}\frac{\mathbf{y}_{-}^{\top}\mathbf{y}_{-}+\frac{1}{n}(\mathbf{y}_{-}^{\top}\mathbf{1}^{n\times n}\mathbf{y}_{-})-\frac{2}{n}(\mathbf{y}_{-}^{\top}\mathbf{1}^{n\times n}\mathbf{y}_{-})}{\mathbf{y}_{-}^{\top}\mathbf{y}_{-}}\\
\nonumber &=&\operatorname*{arg\,min}_{\mathbf{y}_{-}}\frac{\frac{-1}{n}(\mathbf{y}_{-}^{\top}\mathbf{1}^{n\times n}\mathbf{y}_{-})}{\mathbf{y}_{-}^{\top}\mathbf{y}_{-}}\\
&=&\operatorname*{arg\,max}_{\mathbf{y}_{-}}\frac{\mathbf{y}_{-}^{\top}\mathbf{1}^{n\times n}\mathbf{y}_{-}}
{\mathbf{y}_{-}^{\top}\mathbf{y}_{-}}
\label{ray2}
\end{eqnarray}
The relation above is known as the Rayleigh quotient. It is well known that the optimum of the Rayleigh quotient is attained at the eigenvector $\boldsymbol\nu$ corresponding to the largest eigenvalue of the matrix in the numerator. That is, $\mathbf{y}_{-}^{opt}=\boldsymbol\nu$, where in this case $\boldsymbol\nu$ corresponds to the eigenvector corresponding to the largest eigenvalue of $\mathbf{1}^{n\times n}$. It can be easily shown that matrix $\mathbf{1}^{n\times n}$ has a single eigenvector $\boldsymbol\nu$ corresponding to the non-zero eigenvalue of $n$, where $\boldsymbol\nu=(\frac{1}{\sqrt{n}},\dots,\frac{1}{\sqrt{n}})^{\top}$. Note that the Rayleigh quotient is constant under scaling $\mathbf{y}_{-}\rightarrow c\mathbf{y}_{-}$. In other words, if $\mathbf{y}_{-}$ maximises the objective function in Eq. \ref{ray2}, then any non-zero scalar multiple $c\mathbf{y}_{-}$ also maximises Eq. \ref{ray2}. As a result, one may simply choose $\mathbf{y}^{opt}_{-}$ as $\mathbf{y}^{opt}_{-}=(1,\dots,1)^{\top}$ which would lead to $\mathcal{M}=1$.

\subsection{Relation to the Fisher null-space methods}
We now establish the relationship of our formulation in Eq. \ref{ray2} to the null-space Fisher discriminant analysis using the origin as an artificial outlier. For this purpose, first, it is shown that the criterion function in Eq. \ref{ray2} is in fact, the Fisher ratio and then its relation to the null-space approaches is established.

The Fisher analysis maximises the ratio of between-class scatter $S_b$ to the within-class scatter $S_w$. As the negative class is represented by only a single sample (i.e. the origin), it would have a zero scatter and thus the within-class scatter in this case would be $S_w=S(\mathcal{T})$, and hence
\begin{eqnarray}
S_w=\mathbf{y}_{-}^{\top}\mathbf{y}_{-}-\frac{1}{n}\mathbf{y}_{-}^{\top}\mathbf{1}^{n\times n}\mathbf{y}_{-}
\label{wit}
\end{eqnarray}
The between-class scatter when the origin is considered as the mean of the negative class along the direction specified by $f(.)$ is
\begin{eqnarray}
\nonumber S_b&=&(\mathcal{M}-0)^{\top}(\mathcal{M}-0)\\
\nonumber &=&[\frac{1}{n}\sum_{i=1}^{n}f(x_{i})]^2\\
\nonumber &=&\frac{1}{n^2}[(\mathbf{y}_{-}^{\top}\mathbf{1}^{n\times 1})(\mathbf{1}^{1\times n}\mathbf{y}_{-})]\\
&=&\frac{1}{n^2} \mathbf{y}_{-}^{\top}\mathbf{1}^{n\times n}\mathbf{y}_{-}
\label{be1}
\end{eqnarray}
The Fisher analysis maximises the ratio $\frac{S_b}{S_w}$ or equivalently minimises the ratio $\frac{S_w}{S_b}$ and thus
\begin{eqnarray}
\nonumber \mathbf{y}^{opt}_{-} &=& \operatorname*{arg\,min}_{\mathbf{y}_{-}}\frac{S_w}{S_b}=\frac{\mathbf{y}_{-}^{\top}\mathbf{y}_{-}-\frac{1}{n}(\mathbf{y}_{-}^{\top}\mathbf{1}^{n\times n}\mathbf{y}_{-})}{\frac{1}{n^2} \mathbf{y}_{-}^{\top}\mathbf{1}^{n\times n}\mathbf{y}_{-}}\\
\nonumber &=& \operatorname*{arg\,min}_{\mathbf{y}_{-}}[\frac{\mathbf{y}_{-}^{\top}\mathbf{y}_{-}}{\frac{1}{n^2} \mathbf{y}_{-}^{\top}\mathbf{1}^{n\times n}\mathbf{y}_{-}}-\frac{\frac{1}{n}(\mathbf{y}_{-}^{\top}\mathbf{1}^{n\times n}\mathbf{y}_{-})}{\frac{1}{n^2} \mathbf{y}_{-}^{\top}\mathbf{1}^{n\times n}\mathbf{y}_{-}}]\\
\nonumber &=& \operatorname*{arg\,min}_{\mathbf{y}_{-}}[n^2\frac{\mathbf{y}_{-}^{\top}\mathbf{y}_{-}}{\mathbf{y}_{-}^{\top}\mathbf{1}^{n\times n}\mathbf{y}_{-}}]\\
&=& \operatorname*{arg\,max}_{\mathbf{y}_{-}}\frac{\mathbf{y}_{-}^{\top}\mathbf{1}^{n\times n}\mathbf{y}_{-}}{\mathbf{y}_{-}^{\top}\mathbf{y}_{-}}
\label{fisher}
\end{eqnarray}
\noindent which shows that when the negative class is represented by the origin, our criterion function in Eq. \ref{ray2} is in fact the Fisher criterion.

Next, it is shown that the proposed approach is in fact a \textit{null-space} Fisher analysis. The null projection function \cite{8099922,6619277} is defined as a function leading to zero within-class scatter while providing positive between-class scatter. Thus, one needs to show that $\mathbf{y}_{-}^{opt}=(1,\dots,1)^{\top}$ leads to $S_w=0$ and $S_b> 0$. As all the elements of $\mathbf{y}^{opt}_{-}$ are equal, it is clear that the proposed formulation corresponds to a zero scatter for the target class. The conjecture can be also verified by substituting $\mathbf{y}^{opt}_{-}=(1,\dots,1)^{\top}$ in the relation for the within-class scatter as
\begin{eqnarray}
{S_w}|_{\mathbf{y}_{-}^{opt}=(1,\dots,1)^{\top}}=\mathbf{y}_{-}^{\top}\mathbf{y}_{-}-\frac{1}{n}(\mathbf{y}_{-}^{\top}\mathbf{1}^{n\times n}\mathbf{y}_{-})=0
\end{eqnarray}

Next, as all positive training observations are mapped onto point $1$ in the feature subspace while the exemplar outlier is at the origin, the between-class scatter would be $1$. This can be confirmed by substituting $\mathbf{y}^{opt}_{-}=(1,\dots,1)^{\top}$ in the relation for the between-class scatter as:
\begin{eqnarray}
{S_b}|_{\mathbf{y}_{-}^{opt}=(1,\dots,1)^{\top}}=\frac{1}{n^2} \mathbf{y}_{-}^{\top}\mathbf{1}^{n\times n}\mathbf{y}_{-}=1
\end{eqnarray}
As a result, the proposed approach corresponds to a projection function (i.e. $f(.)$) leading to $S_w=0$ and $S_b=1$ and hence is a null-space Fisher analysis similar to \cite{8099922,6619277}.
\subsection{Extension to the supervised case: the use of counter-examples}
Up to this point, it is assumed that the training data solely consists of positive samples. Although in a one-class classification problem negative samples are not expected to be available abundantly, nevertheless, in case some negative observations exist, they may be used to refine the solution. In this section, the proposed method is extended to benefit from the availability of labelled non-target observations in the training set in a supervised operating mode. Recall that we are minimising the scatter of training samples while keeping the projections of positive samples distant from the centre of the negative class (i.e. the origin) in the feature space. As a result, if some negative samples are available, first, the centre of the negative class needs to be shifted to zero to be consistent with our earlier assumption of having the mean of negative class located at the origin. Next, in the computation of the total scatter, both positive and negative training observations shall be included. These two modifications to the problem are discussed next.

Without loss of generality, let us assume that the last $n_0$ samples in the training set $\{x_i\in\mathbb{R}^d|i=1,\dots, n\}$ correspond to negative training observations. The mean of the non-target training samples is thus
\begin{eqnarray}
\mathcal{O}=\frac{1}{n_0}(\overbrace{0,\dots,0}^{n-n_0},\overbrace{1,\dots,1}^{n_0})\mathbf{y}_{-}
\end{eqnarray}

Next, let us define the $(n-n_0)\times n$ matrix $\mathbf{T_1}$ as a concatenation of an $(n-n_0)\times (n-n_o)$ identity matrix and an $(n-n_0)\times n_0$ zero matrix as
\begin{eqnarray}
 \mathbf{T_1}=\left( \begin{matrix} 
	1 &0  & \dots  &0 & | &0&\dots & 0\\
	0 &1  &\dots & \vdots&|& 0& \dots & 0\\
	\vdots &\vdots &1 & 0 &|& \vdots &  &\vdots\\
	0& \dots & 0& 1& |&0& \dots & 0
   \end{matrix}\right)
	\label{matT1}
\end{eqnarray}

Using $\mathcal{O}$ and $\mathbf{T_1}$, the transformed positive samples, denoted as $\mathbf{y}_{-tp}$, which correspond to the positive samples of the training set shifted by the mean of the negative training samples can be written as
\begin{eqnarray}
\nonumber \mathbf{y}_{-tp}&=&\mathbf{T_1}\mathbf{y}_{-}-\frac{1}{n_0}\mathbf{1}^{(n-n_0)\times 1}(\overbrace{0,\dots,0}^{n-n_0},\overbrace{1,\dots,1}^{n_0}) \mathbf{y}_{-}\\
\nonumber &=& (\mathbf{T_1}-\frac{1}{n_0}\mathbf{1}^{(n-n_0)\times 1}(\overbrace{0,\dots,0}^{n-n_0},\overbrace{1,\dots,1}^{n_0})) \mathbf{y}_{-}\\
&=& \mathbf{G_1}\mathbf{y}_{-}
\end{eqnarray}
\noindent where $\mathbf{G_1}=\mathbf{T_1}-\frac{1}{n_0}\mathbf{1}^{(n-n_0)\times 1}(\overbrace{0,\dots,0}^{n-n_0},\overbrace{1,\dots,1}^{n_0})$.
Let us also define the $n_0\times n$ matrix $\mathbf{T_2}$ as a concatenation of an $n_0\times (n-n_o)$ zero matrix and an $n_0\times n_0$ identity matrix as
\begin{eqnarray}
 \mathbf{T_2}=\left( \begin{matrix} 
	0&\dots & 0 &| &1 &0  & \dots  &0 & \\
	0& \dots & 0 &| & 0 &1  &\dots & \vdots&\\
	 \vdots &  &\vdots &| & \vdots &\vdots &1 & 0\\
	0& \dots & 0 &| & 0& \dots & 0& 1
   \end{matrix}\right)
	\label{matT2}
\end{eqnarray}
Using $\mathcal{O}$ and $\mathbf{T_2}$, transformed negative samples, denoted as $\mathbf{y}_{-tn}$, which correspond to the negative samples of the training set shifted by the mean of the negative training samples can be written as
\begin{eqnarray}
\nonumber \mathbf{y}_{-tn}&=&\mathbf{T_2}\mathbf{y}_{-}-\frac{1}{n_0}\mathbf{1}^{n_0\times 1}(\overbrace{0,\dots,0}^{n-n_0},\overbrace{1,\dots,1}^{n_0}) \mathbf{y}_{-}\\
\nonumber &=& (\mathbf{T_2}-\frac{1}{n_0}\mathbf{1}^{n_0\times 1}(\overbrace{0,\dots,0}^{n-n_0},\overbrace{1,\dots,1}^{n_0})) \mathbf{y}_{-}\\
&=& \mathbf{G_2}\mathbf{y}_{-}
\end{eqnarray}
\noindent where $\mathbf{G_2}=\mathbf{T_2}-\frac{1}{n_0}\mathbf{1}^{n_0\times 1}(\overbrace{0,\dots,0}^{n-n_0},\overbrace{1,\dots,1}^{n_0})$. Note that $\mathbf{y}_{-tp}$ and $\mathbf{y}_{-tn}$ correspond to the projections of positive and negative samples onto the feature space with the additional property that the mean of the negative set again lies at the origin. The within-class scatter in this case would be the sum of scatters corresponding to the positive and negative sets, i.e. $S_w=S_p+S_n$. Considering Eq. \ref{wit}, $S_p$ is now given as
\begin{eqnarray}
\nonumber S_p&=&\mathbf{y}_{-tp}^{\top}\mathbf{y}_{-tp}-\frac{1}{n-n_0}\mathbf{y}_{-tp}^{\top}\mathbf{1}^{(n-n_0)\times (n-n_0)}\mathbf{y}_{-tp}\\
\nonumber &=&\mathbf{y}_{-}^{\top}\mathbf{G_1}^{\top}\mathbf{G_1}\mathbf{y}_{-}-\frac{1}{n-n_0}\mathbf{y}_{-}^{\top}\mathbf{G_1}^{\top}\mathbf{1}^{(n-n_0)\times (n-n_0)}\mathbf{G_1}\mathbf{y}_{-}
\end{eqnarray}
and similarly, $S_n$ is given as
\begin{eqnarray}
\nonumber S_n&=&\mathbf{y}_{-tn}^{\top}\mathbf{y}_{-tn}-\frac{1}{n_0}\mathbf{y}_{-tn}^{\top}\mathbf{1}^{n_0\times n_0}\mathbf{y}_{-tn}\\
\nonumber &= &\mathbf{y}_{-}^{\top}\mathbf{G_2}^{\top}\mathbf{G_2}\mathbf{y}_{-}-\frac{1}{n_0}\mathbf{y}_{-}^{\top}\mathbf{G_2}^{\top}\mathbf{1}^{n_0\times n_0}\mathbf{G_2}\mathbf{y}_{-}
\end{eqnarray}

Since the mean of the transformed negative set is located at the origin, drawing on Eq. \ref{be1}, the between-class scatter is given as
\begin{eqnarray}
\nonumber S_b&=&\frac{1}{(n-n_0)^2} \mathbf{y}_{-tp}^{\top}\mathbf{1}^{(n-n_0)\times (n-n_0)}\mathbf{y}_{-tp}\\
&=&\frac{1}{(n-n_0)^2} \mathbf{y}_{-}^{\top}\mathbf{G_1}^{\top} \mathbf{1}^{(n-n_0)\times (n-n_0)} \mathbf{G_1}\mathbf{y}_{-}
\end{eqnarray}

Minimising scatter while maximising average margin between projections of target observations and the origin would then lead to
\begin{IEEEeqnarray}{l}
\nonumber \mathbf{y}_{-}^{opt} = \operatorname*{arg\,min}_{\mathbf{y}_{-}} \frac{S_p+S_n}{S_b}\\
\nonumber = \operatorname*{arg\,min}_{\mathbf{y}_{-}} [\frac{\mathbf{y}_{-}^{\top}\mathbf{G_1}^{\top}\mathbf{G_1}\mathbf{y}_{-}}{\mathbf{y}_{-}^{\top}\mathbf{G_1}^{\top} \mathbf{1}^{(n-n_0)\times (n-n_0)} \mathbf{G_1}\mathbf{y}_{-}}\\
\nonumber -\frac{\mathbf{y}_{-}^{\top}\mathbf{G_1}^{\top}\mathbf{1}^{(n-n_0)\times (n-n_0)}\mathbf{G_1}\mathbf{y}_{-}}{\mathbf{y}_{-}^{\top}\mathbf{G_1}^{\top} \mathbf{1}^{(n-n_0)\times (n-n_0)} \mathbf{G_1}\mathbf{y}_{-}}\\
\nonumber+\frac{\mathbf{y}_{-}^{\top}\mathbf{G_2}^{\top}\mathbf{G_2}\mathbf{y}_{-}}{\mathbf{y}_{-}^{\top}\mathbf{G_1}^{\top} \mathbf{1}^{(n-n_0)\times (n-n_0)} \mathbf{G_1}\mathbf{y}_{-}}\\
-\frac{\mathbf{y}_{-}^{\top}\mathbf{G_2}^{\top}\mathbf{1}^{n_0 \times n_0}\mathbf{G_2}\mathbf{y}_{-}}{\mathbf{y}_{-}^{\top}\mathbf{G_1}^{\top} \mathbf{1}^{(n-n_0)\times (n-n_0)} \mathbf{G_1}\mathbf{y}_{-}}]
\label{tp}
\end{IEEEeqnarray}
\noindent It can be easily verified that the last term in Eq. \ref{tp} corresponds to a scalar multiple of the mean of shifted negative examples and hence is zero and subsequently $\mathbf{y}_{-}^{opt}$ is given as
\begin{IEEEeqnarray}{l}
\nonumber \mathbf{y}_{-}^{opt} = \operatorname*{arg\,min}_{\mathbf{y}_{-}}\frac{\mathbf{y}_{-}^{\top}(\mathbf{G_1}^{\top}\mathbf{G_1}+\mathbf{G_2}^{\top}\mathbf{G_2})\mathbf{y}_{-}}{\mathbf{y}_{-}^{\top}\mathbf{G_1}^{\top} \mathbf{1}^{(n-n_0)\times (n-n_0)} \mathbf{G_1}\mathbf{y}_{-}}\\
= \operatorname*{arg\,max}_{\mathbf{y}_{-}}\frac{\mathbf{y}_{-}^{\top}\mathbf{G_1}^{\top} \mathbf{1}^{(n-n_0)\times (n-n_0)} \mathbf{G_1}\mathbf{y}_{-}}{\mathbf{y}_{-}^{\top}(\mathbf{G_1}^{\top}\mathbf{G_1}+\mathbf{G_2}^{\top}\mathbf{G_2})\mathbf{y}_{-}}
\label{tp2}
\end{IEEEeqnarray}
Eq. \ref{tp2} is a generalised Rayleigh quotient the solution of which is given by the generalised eigen-value problem
\begin{eqnarray}
\mathbf{G_1}^{\top}\mathbf{1}^{(n-n_0)\times(n-n_0)}\mathbf{G_1}\mathbf{y}_{-}^{opt}=\lambda (\mathbf{G_1}^{\top}\mathbf{G_1}+\mathbf{G_2}^{\top}\mathbf{G_2}) \mathbf{y}_{-}^{opt}
\end{eqnarray}
The eigenvector corresponding to the maximum eigenvalue satisfying the problem above is
\begin{eqnarray}
\mathbf{y}_{-}^{opt}=(\overbrace{1,\dots,1}^{n-n_o},\overbrace{0,\dots,0}^{n_0})^{\top}
\label{opts}
\end{eqnarray}
However, any solution of the generic form $\mathbf{y}_{-}^{opt}=(\overbrace{c_1,\dots,c_1}^{n-n_o},\overbrace{c_2,\dots,c_2}^{n_0})^{\top}$ (s.t. $c_1\neq c_2$) would result in a zero within-class scatter while providing a positive between-class scatter and is equally applicable.

\subsection{Spectral Regression}
Once $\mathbf{y}_{-}^{opt}$ is determined, the relation $\mathbf{y}_{-}^{opt}=\mathbf{K}^{\top}\boldsymbol\alpha^{opt}$ may be used to determine $\boldsymbol\alpha^{opt}$. This approach is called \textit{spectral regression} in \cite{SRKDA}. Rewriting Eq. \ref{tp2} in a general form as
\begin{IEEEeqnarray}{l}
\mathbf{y}_{-}^{opt} = \operatorname*{arg\,min}_{\mathbf{y}_{-}} \frac{\mathbf{y}_{-}^{\top}\mathbf{W}\mathbf{y}_{-}}{\mathbf{y}_{-}^{\top}\mathbf{Q}\mathbf{y}_{-}}
\end{IEEEeqnarray}
\noindent The spectral regression approach involves two steps to solve for $\boldsymbol{\alpha}$:
\begin{enumerate}
\item Solve $\mathbf{W}\boldsymbol{\nu}=\lambda \mathbf{Q} \boldsymbol{\nu}$  for $\boldsymbol{\nu}$;
\item Solve $\mathbf{K}\boldsymbol{\alpha} = \boldsymbol{\nu}$  for $\boldsymbol{\alpha}$.
\end{enumerate}
The method is dubbed \textit{spectral regression} as it involves spectral analysis for the problem $\mathbf{W}\boldsymbol{\nu}=\lambda \mathbf{Q} \boldsymbol{\nu}$ followed by solving $\mathbf{K}^{\top}\boldsymbol{\alpha} = \boldsymbol{\nu}$ which is equivalent to a regularised regression problem \cite{SRKDA}. However, in our formulation, due to the special structures of $\mathbf{W}$ and $\mathbf{Q}$, the leading eigenvector could be directly found.

Solving $\mathbf{y}^{opt}_{-}=\mathbf{K}^{\top}\boldsymbol{\alpha}^{opt}$ for $\boldsymbol{\alpha}^{opt}$ can be performed using the Cholesky factorisation and forward-back substitution. In this case, if $\mathbf{K}$ is positive-definite, then there exists a unique solution for $\boldsymbol{\alpha}$. If $\mathbf{K}$ is singular, it is approximated by the positive definite matrix $\mathbf{K}+\delta \mathbf{I}$ where $\mathbf{I}$ is the identity matrix and $\delta > 0$ is a regularisation parameter. As a widely used kernel function, the radial basis kernel function, {\em i.e.} $K_{ij}=\kappa(x_i,x_j)=e^{-\Vert\mathbf{x}_i-\mathbf{x}_j\Vert^2/{2\sigma^2}}$, leads to a positive definite kernel matrix \cite{SRKDA,kernelbook} for which $\delta=0$ and the spectral regression finds the exact solution. 
Considering a Cholesky factorisation of $\mathbf{K}$ as $\mathbf{K}=\mathbf{R}^\top \mathbf{R}$, $\boldsymbol{\alpha}$ may be found by first solving $\mathbf{R}^\top\boldsymbol{\theta}=\boldsymbol{\nu}\nonumber$ for $\boldsymbol{\theta}$ and then solving $\mathbf{R}\boldsymbol{\alpha}=\boldsymbol{\boldsymbol{\theta}}$ for $\boldsymbol{\alpha}$. Since in the proposed approach there is only one eigenvector associated with the equation $\mathbf{K}\boldsymbol{\alpha} = \boldsymbol{\nu}$, only a \textit{single} vector, {\em i.e.} $\boldsymbol{\alpha}^{opt}$, is computed.
\subsection{Outlier Detection}
Once $\boldsymbol\alpha^{opt}$ is determined, the projection of a probe $\mathbf{z}$ onto the optimal feature subspace can be obtained as $f(\mathbf{z})=\sum_{i=1}^{n}\alpha^{opt}_i\kappa(\mathbf{z},\mathbf{x}_i)=\mathbf{k}_z^\top\boldsymbol{\alpha}^{opt}$, where $\mathbf{k}_z$ is a vector collection of the elements $\kappa(\mathbf{z},\mathbf{x}_i)$. The decision rule is now defined as the distance between the mean of projections of positive training observations in the feature space, i.e. $\mathcal{M}$ and $f(\mathbf{z})$. As $\mathcal{M}=1$, the decision rule becomes
\begin{eqnarray}
|\mathbf{k}_z^\top\boldsymbol{\alpha}^{opt}-1|  > \tau& \hspace{1cm} &\mbox{\textit{z is an outlier}}\nonumber \\ 
|\mathbf{k}_z^\top\boldsymbol{\alpha}^{opt}-1|  \leq \tau& \hspace{1cm} &\mbox{\textit{z is a target object}}
\label{dr}
\end{eqnarray}
where $\tau$ is a threshold for deciding normality. Two observations regarding the decision rule and the decision threshold are in order. First, as the Rayleigh quotient is constant under scaling, the value of $\mathbf{y}^{opt}$ (and consequently $\mathcal{M}$) can be chosen arbitrarily as long as all the elements of $\mathbf{y}^{opt}$ are equal. This freedom is reflected in the decision rule as choosing a different $\mathbf{y}^{opt}$ (leading to a different $\mathcal{M}$ other than $1$) would only introduce a scaling (e.g. $c$) on $\boldsymbol\alpha^{opt}$ due to the relation $\mathbf{K}\boldsymbol\alpha^{opt}=\mathbf{y}^{opt}$. In this case, the same scaling effect (i.e. $c$) would be applied to $f(\mathbf{z})=\mathbf{k}_z^\top\boldsymbol{\alpha}^{opt}$. As a result, the same decision rule in Eq. \ref{dr} would be still valid by using the new threshold $\tau/c$. In other words, the choice of a particular value for the elements of $\mathbf{y}^{opt}$ only introduces a scaling effect on the threshold and does not affect the performance as long as numerical errors due to finite precision of computations do not occur. Second, since there is only one single point on the feature subspace corresponding to the projection of positive training samples and another single point corresponding to the projection of negative training instances, a threshold can be set to reject all negative training samples while accepting all positive training observations. However, finding a threshold to reject an \textit{arbitrary} proportion of training samples when they are all utilised for training is not feasible. Nevertheless, if an arbitrary proportion of the training data shall be rejected, a leave-one-out training scheme on the training set may be followed to produce $n$ possibly distinct scores for the training samples on which a threshold may be set experimentally to reject a desired proportion of the data.

The pseudo-codes corresponding to the training and testing stages of the proposed OC-KSR approach are summarised in the Algorithms \ref{OC-KSR-train} and \ref{OC-KSR-test}.
\begin{algorithm}
\footnotesize
\caption{Training}\label{OC-KSR-train}
\begin{algorithmic}[1]
\State \texttt{Set} $n = \#$ \texttt{total samples} \& $n_0 = \#$ \texttt{negative examples}
\State \texttt{Set} $\mathbf{\nu}=(\overbrace{1,\dots,1}^{n-n_o},\overbrace{0,\dots,0}^{n_0})$
\State \texttt{Calculate} $\mathbf{K}$
\State \texttt{Form the Cholesky decomposition of} $\mathbf{K}$: \begin{equation}
\mathbf{K}= \mathbf{R}^\top \mathbf{R}
\label{Km}\nonumber
\end{equation}
\State \texttt{Solve} $\mathbf{R}^\top\boldsymbol{\theta}=\boldsymbol{\nu}\nonumber$ \texttt{for} $\boldsymbol{\theta}$
\State \texttt{Solve} $\mathbf{R}\boldsymbol{\alpha}=\boldsymbol{\theta}$ \texttt{for} $\boldsymbol{\alpha}$
\State \texttt{output} $\boldsymbol{\alpha}$
\end{algorithmic}
\normalsize
\end{algorithm}

\begin{algorithm}[t]
\footnotesize
\caption{Testing probe $z$}\label{OC-KSR-test}
\begin{algorithmic}[1]
\State \texttt{compute} $\mathbf{k}_{z}=[\kappa(x_1,z),\dots ,\kappa(x_{n},z)]^\top$
\State \texttt{compute} $f(z)=  \mathbf{k}_z^\top\boldsymbol{\alpha}^{opt}$
\If{$ |f(z)-1| \leq \tau$}
 \State $z$ \texttt{is a target object}
 \Else
\State $z$ \texttt{is an outlier}
\EndIf
\end{algorithmic}
\normalsize
\end{algorithm}

\subsection{Incremental OC-KSR}
In the proposed OC-KSR method, a high computational cost is associated with the Cholesky decomposition of the kernel matrix $\mathbf{K}$, the batch computation of which requires $O(n^3)$ arithmetic operations. However, as advocated in \cite{matrixdecom}, a Cholesky decomposition may be obtained more efficiently using an incremental approach. In the incremental scheme, the goal is to find the Cholesky decomposition of an $m\times m$ matrix given the Cholesky decomposition of its $(m-1)\times (m-1)$ submatrix. Hence, given the Cholesky decomposition of the kernel matrix $\mathbf{K}^{(m-1)\times (m-1)}$ of $m-1$ samples, one is interested in computing the Cholesky factorisation of the kernel matrix $\mathbf{K}^{m\times m}$ for the augmented training set where a single sample ($x_m$) is injected into the system. The incremental Cholesky decomposition technique may be applied via the \textit{Sherman's March} algorithm \cite{matrixdecom} for $\mathbf{K}^{m\times m}$ as
\begin{IEEEeqnarray}{l}
\nonumber \mathbf{K}^{m\times m}=\\
\nonumber \begin{pmatrix}
  \mathbf{K}^{(m-1)\times (m-1)} & \mathbf{k}_{1m} \\
	\mathbf{k}_{1m}^\top & k_{mm}
\end{pmatrix}=\\
\begin{pmatrix}
  \mathbf{R}^{(m-1)\times (m-1)^\top} & \mathbf{0} \\
	\mathbf{r}_{1m}^\top & r_{mm}
\end{pmatrix}
\begin{pmatrix}
  \mathbf{R}^{(m-1)\times(m-1)} & \mathbf{r}_{1m} \\
	\mathbf{0} & r_{mm}	
\end{pmatrix}
\label{Km1}
\end{IEEEeqnarray}
where $\mathbf{k}_{1m}$ is an $(m-1)\times 1$ vector given by\\ $\mathbf{k}_{1m}=[\kappa(x_1,x_m),\dots ,\kappa(x_{m-1},x_m)]^\top$ and \\$k_{mm}=\kappa(x_m,x_m)$.\\Eq. \ref{Km1} reads
\begin{eqnarray}
\mathbf{K}^{(m-1)\times(m-1)}=\mathbf{R}^{(m-1)\times(m-1)^\top} \mathbf{R}^{(m-1)\times(m-1)} \nonumber \\
\mathbf{k}_{1m} = {\mathbf{R}_{m-1}^\top} \mathbf{r}_{1m}\nonumber \\
 r_{mm} = \sqrt{k_{mm}-\mathbf{r}_{1m}^\top \mathbf{r}_{1m}}
\end{eqnarray}
Thus, one first solves $\mathbf{k}_{1m} = {\mathbf{R}^{(m-1)\times(m-1)^\top}} \mathbf{r}_{1m}$ for $\mathbf{r}_{1m}$ and then computes $r_{mm}$. The employed incremental technique reduces the computational cost of the Cholesky decomposition from cubic in number of training samples in the batch mode to quadratic in the incremental mode \cite{SRKDA}.

By varying $m$ from 1 to $n$ (total number of target observations), the incremental Cholesky decomposition is derived as Algorithm \ref{inc_chol}. The incremental approach not only reduces the computational complexity but also allows for operation in streaming data scenarios. In this case, as new data becomes available, only the new part of the kernel matrix $\mathbf{K}$ needs to be computed. Moreover, since the Cholesky factorisation can be performed in an incremental fashion, the previous computations are fully utilised.

\begin{algorithm}[t]
\footnotesize
\caption{Incremental Cholesky decomposition}\label{inc_chol}
\begin{algorithmic}[1]
\State \texttt{Set} $\mathbf{R}^{0}=\sqrt{\kappa(x_1,x_1)}$
\For{\texttt{ m=2:n}}
 \State $\mathbf{k}_{1m}=[\kappa(x_1,x_m),\dots ,\kappa(x_{m-1},x_m)]^\top$
 \State \texttt{Find} $\mathbf{r}_{1m}$ \texttt{satisfying} $\mathbf{k}_{1m} = {\mathbf{R}^{(m-1)\times(m-1)^\top}} \mathbf{r}_{1m}$
 \State $k_{mm}=\kappa(x_m,x_m)$
 \State $r_{mm} = \sqrt{k_{mm}-\mathbf{r}_{1m}^\top \mathbf{r}_{1m}}$
 \State $\mathbf{R}^{(m-1)\times(m-1)}=\begin{pmatrix}
  \mathbf{R}^{(m-1)\times(m-1)} & \mathbf{r}_{1m} \\
	\mathbf{0} & r_{mm}
\end{pmatrix}$
\EndFor
 \State \texttt{output} $\mathbf{R}=\mathbf{R}_{m-1}$
\end{algorithmic}
\normalsize
\end{algorithm}

\subsection{Discussion}
\label{difs}
There exist some unsupervised methods using the kernel PCA (KPCA) approach for outlier detection such as those in \cite{HOFFMANN2007863,GueSchVis07}. If in the KPCA approach one uses the eigenvector corresponding to the smallest eigenvalue for projection, a small variance along the projection direction is expected. Note that in KPCA one may obtain at most $n$ ($n$ being the number of training samples) distinct eigenvalues using the kernel matrix. As the smallest eigenvalue of a general kernel matrix need not be zero, the variance along the corresponding eigenvector would not necessarily be zero. As a widely used kernel function, an RBF kernel results in a positive-definite kernel matrix which translates into strictly positive eigenvalues. In contrast, in the proposed method, the variance along the projection direction is zero even when using an RBF kernel function.

As discussed previously, the proposed method is similar to the null-space methods for novelty detection presented in \cite{8099922,6619277} in the sense that all methods employ the Fisher criterion for estimation of a null feature space. However, the proposed approach, as will be discussed in \S \ref{comps} is computationally attractive by virtue of avoiding costly eigen-decompositions. Other work in \cite{7727608} tries to optimise the ratio between the target scatter and outlier scatter which is different from the Fisher ratio utilised in this work. As illustrated, the proposed approach can be implemented in an incremental fashion which further reduces the computational complexity of the method while allowing for application in streaming data scenarios. Moreover, the proposed OC-KSR method can employ possible labelled negative training observations to refine the decision boundary.

\section{Experimental Evaluation}
\label{exps}
In this section, an experimental evaluation of the proposed approach is provided to compare the performance of the OC-KSR method to those of several
state-of-the-art approaches in terms of the area under the ROC curve (AUC). Ten different datasets which include relatively low to medium and high dimensional feature sets are used for this purpose. A summary of the statistics of the datasets used is provided in Table \ref{Characteristics} where $d$ denotes the dimensionality of feature sets. A brief description regarding the datasets used in the experiments is as follows.
\begin{itemize}
\item \textbf{Arcene}: The task in this dataset is to distinguish cancer versus normal patterns from mass-spectrometric data. The dataset was obtained by merging three mass-spectrometry datasets with continuous input variables to obtain training and test data. The dataset is part of the 2003 NIPS variable selection benchmark. The original features indicate the abundance of proteins in human sera having a given mass value. Based on these features, one must separate cancer patients from healthy patients. The dataset is part of the UCI machine learning datasets \cite{UCIwebsite}.
\item \textbf{AD} includes EEG signals from 11 patients with a diagnosis of a probable Alzheimer's Disease (AD) and 11 controls subjects. The task in this dataset is to discriminate healthy subjects from AD patients. AD patients were recruited from the Alzheimer's Patients' Relatives Association of Valladolid (AFAVA), Spain for whom more than 5 minutes of EEG data were recorded using Oxford Instruments Profile Study Room 2.3.411 (Oxford, UK) \cite{8096133}. As suggested in \cite{8096133}, in this work the signal associated with the $O_2$ electrode is used.
\item \textbf{Face} consists of face images of different individuals where the task is to recognise a subject among others. For each subject, a one-class classifier is built using the data associated with that subject while all other subjects are considered as outliers with respect to the built model. The experiment is repeated in turn for all subjects in the dataset. The features used for image representation are obtained via the GoogleNet deep CNN \cite{7298594}. We have created this dataset out of the real-access data of the Replay-Mobile dataset \cite{Costa-Pazo_BIOSIG2016_2016} and included ten subjects in the experiments.
\item \textbf{Caltech256} is a challenging set of 256 object categories containing 30607 images in total \cite{griffinHolubPerona}. Each class of images has a minimum of 80 images representing a diverse set of backgrounds, poses, lighting conditions and image sizes. In this experiment, the 'American-flag' is considered as the target class and the samples associated with the 'boom-box', 'bulldozer' and 'cannon' classes as outliers. Bag-of-visual-words histograms from densely sampled SIFT features are used to represent images \footnote{http://homes.esat.kuleuven.be/ \~ tuytelaa/unsup \_ features.html}.
\item \textbf{MNIST} is a collection of $28\times 28$ pixel images of handwritten digits 0-9 \cite{726791}. Considering digit '1' as the target digit, 220 images are used as target data and 293 images corresponding to other digits are used as negative samples. Raw image intensities are used for the experiments on this dataset.
\item \textbf{Delft pump} includes 5 vibration measurements taken under different normal and abnormal conditions from a submersible pump. The 5 measurements are combined into one object, giving a 160-dimensional feature space. The dataset is obtained from the one-class dataset archive of Delft university \cite{delftwebsite}.
\item \textbf{Sonar} is composed of 208 instances of 60 attributes representing the energy within a particular frequency band, integrated over a certain period of time. There are two classes: an object is a rock or is a mine. The task is to discriminate between sonar signals bounced off a metal cylinder and those bounced off a roughly cylindrical rock. The Sonar dataset is from the undocumented databases from UCI.
\item \textbf{Vehicle} dataset is from Statlog, where the class van is used as a target class. The task is to recognise a vehicle from its silhouette. The dataset is obtained from the one-class dataset archive of Delft university \cite{delftwebsite}.
\item \textbf{Vowel} is an undocumented dataset from UCI. The purpose is speaker independent recognition of the eleven steady state vowels of British English using a specified training set of lpc derived log area ratios. Vowel 0 is used as the target class in this work.
\item \textbf{Balance-scale} was generated to model psychological experimental results. Each example is classified as having the balance scale tip to the right, tip to the left, or be balanced. The attributes are the left weight, the left distance, the right weight, and the right distance. The dataset is part of the UCI machine learning repository \cite{UCIwebsite}.
\end{itemize}

The methods included in the comparison are as follows:
\begin{itemize}
\item \textbf{OC-KSR} is the proposed one-class spectral regression when negative training samples are not present in the training set.
\item \textbf{SVDD} is the Support Vector Data Description approach to solve the one class classification problem \cite{Tax2004}. As a widely used method, it provides a baseline for comparison.
\item \textbf{OC-KNFST} The one-class kernel null Foley-Sammon transform presented in \cite{6619277} which operates on the Fisher criterion. This method is chosen due to its similarity to the proposed approach.
\item \textbf{KPCA} is based on the kernel PCA method where the reconstruction residual of a sample in the feature space is used as the novelty measure \cite{HOFFMANN2007863}.
\item \textbf{GP} is derived based on the Gaussian process regression and approximate Gaussian process classification \cite{KEMMLER20133507} where in this work the predictive mean is used as one class score.
\item \textbf{LOF} Local outlier factor (LOF) \cite{Breunig00lof:identifying} is a local measure indicating the degree of novelty for each object of the dataset. The LOF of an object is based on a single parameter k, which is the number of nearest neighbours used in defining the local neighbourhood of the object.
\item \textbf{K-means} is the k-means clustering based approach where k centres are assumed for the target observation. The novelty score of a sample is defined as the minimum distance of a query to data centres.
\item \textbf{KNNDD} The k-nearest neighbours data description method (KNNDD) is proposed in terms of the one class classification framework \cite{30e27ea86}. The principle of KNDD is to associate to each data a distance measure relative to its neighbourhood (k-neighbours).
\end{itemize}
In all the experiments that follow, the positive samples of each dataset are divided into training and test sets of equal sizes randomly. Each experiment is repeated 100 times using random splits of data and the average area under the ROC curve (AUC) and the standard deviation of the AUC's are reported. Furthermore, a statistical analysis is performed to derive average relative rankings of different approaches \cite{DERRAC20113}. No pre-processing of features is performed other than normalising all features to have a unit L2-norm. For the methods requiring a neighbourhood parameter (i.e. LOF, K-means and KNDD), the neighbourhood parameter is set in the range $[3,\dots,10]$ to obtain the best performance. Regarding the methods operating in the RKHS space (i.e. SVDD, OC-KNFST, GP, KPCA and OC-KSR), a common Gaussian kernel is computed and shared among all methods.

\begin{table}
\renewcommand{\arraystretch}{1.2}
\caption{Characteristics of different datasets\newline ($\mathbf{\lowercase{d}}$ denotes dimensionality)}
\label{Characteristics}
\centering
\begin{tabular}{lccc}
\hline
\textbf{Dataset} & $\boldsymbol{\#}$ \textbf{Positive Instances} & $\boldsymbol{\#}$ \textbf{Negative Instances} & \textbf{d}\\
\hline
Arcene	&88	&112	&10000\\
AD	&263	&400	&1280\\
Face	&10$\times$290	&10$\times$290	&1024\\
Caltech256	&97	&304	&1000\\
MNIST	&220	&293&	784\\
Pump	&189	&531&	160\\
Sonar&111	&97	&60\\
Vehicle	&199&	647&	18\\
Vowel	&48&	480	&10\\
Balance-scale	&49	&576&	4\\
\hline
\end{tabular}
\end{table}

\subsection{Comparison to other methods}
A comparison of the proposed OC-KSR approach to other methods is provided in Tables \ref{AUCsHigh} and \ref{AUCsLow} for the datasets with medium to high dimensional features and datasets with relatively lower dimensional features, respectively. From Tables \ref{AUCsHigh} and \ref{AUCsLow}, one may observe that in 4 out of 10 datasets, the proposed OC-KSR method achieves leading performance and on 3 others is placed second in terms of average AUC. Tables \ref{Sum1}, \ref{Sum2} and \ref{Sum3} report the results of a statistical analysis for the significance of the results using the Friedman Test \cite{DERRAC20113} to infer average rankings of different methods. As can observed from Table \ref{Sum1}, the best performing methods on the medium to high dimensional datasets in terms of average ranking are the proposed OC-KSR and the OC-KNFST method, closely followed by AVDD and KPCA. It is worth noting that the performances of both OC-KSR and the OC-KNFST methods do exactly match. As previously discussed, this is expected since both approaches are equivalent theoretically, optimising the Fisher criterion for classification. 

Regarding the lower dimensional datasets, the best-performing methods in terms of average ranking are the proposed OC-KSR approach and the OC-KNFST method, Table \ref{Sum2}. The second best performing method, is GP followed by KPCA.

Table \ref{Sum3} reports the average rankings for all the evaluated methods over all datasets regardless of the dimensionality of feature vectors. The best performing methods are those of OC-KSR and OC-KNFT while the next best performing methods are KPCA and SVDD.
\begin{table*}
\renewcommand{\arraystretch}{1.2}
\caption{Mean AUC’s (+- std) ($\%$) over 100 repetitions on datasets with medium to high dimensional feature vectors}
\label{AUCsHigh}
\centering
\begin{tabular}{l}
\includegraphics[scale=.6]{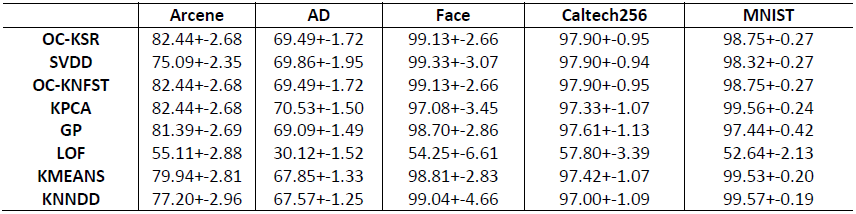}\\
\end{tabular}
\end{table*}

\begin{table*}
\renewcommand{\arraystretch}{1.2}
\caption{Mean AUC’s (+- std) ($\%$) over 100 repetitions on datasets with relatively lower dimensional feature vectors}
\label{AUCsLow}
\centering
\begin{tabular}{l}
\includegraphics[scale=.6]{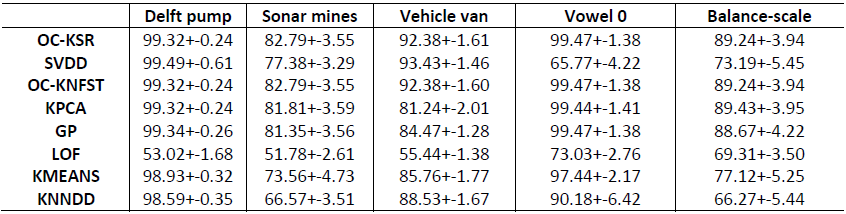}\\
\end{tabular}
\end{table*}

\begin{table}
\renewcommand{\arraystretch}{1.2}
\caption{Average rankings of different methods on data sets with medium to high dimensional feature vectors (Friedman)-Friedman statistic distributed according to chi-square with 7 degrees of freedom: 16.516666666666637. P-value computed by Friedman Test: 0.02079321144566737.}
\label{Sum1}
\centering
\begin{tabular}{lc}
\hline
Method & Ranking\\
\hline
OC-KSR (this work) & 2.9\\
SVDD & 3.59\\
OC-KNFST & 2.9\\
KPCA & 3.6\\
GP & 5.19\\
LOF & 8.0\\
KMEANS & 4.8\\
KNNDD & 4.99\\
\hline
\end{tabular}
\end{table}

\begin{table}
\renewcommand{\arraystretch}{1.2}
\caption{Average rankings of different methods on data sets with relatively lower dimensional feature vectors(Friedman)-Friedman statistic distributed according to chi-square with 7 degrees of freedom: 19.516666666666705. P-value computed by Friedman Test: 0.006713973112715599.}
\label{Sum2}
\centering
\begin{tabular}{lc}
\hline
Method & Ranking\\
\hline
OC-KSR (this work)&	2.5\\
SVDD& 4.2\\
OC-KNFST& 2.5\\
KPCA& 3.8\\
GP& 3.6\\
LOF& 7.6\\
KMEANS& 5.4\\
KNNDD& 6.4\\
\hline
\end{tabular}
\end{table}

\begin{table}
\renewcommand{\arraystretch}{1.2}
\caption{Average rankings of different methods on all datasets (Friedman)-Friedman statistic distributed according to chi-square with 7 degrees of freedom: 33.63333333333335. P-value computed by Friedman Test: 2.017014429267494E-5.}
\label{Sum3}
\centering
\begin{tabular}{lc}
\hline
Method & Ranking\\
\hline
OC-KSR (this work)&	2.7\\
SVDD	& 3.9\\
OC-KNFST& 2.7\\
KPCA& 3.7\\
GP& 4.4\\
LOF& 7.8\\
KMEANS& 5.1\\
KNNDD& 5.69\\
\hline
\end{tabular}
\end{table}

\subsection{Training sample size}
\begin{figure}
\center
\includegraphics[scale=.36]{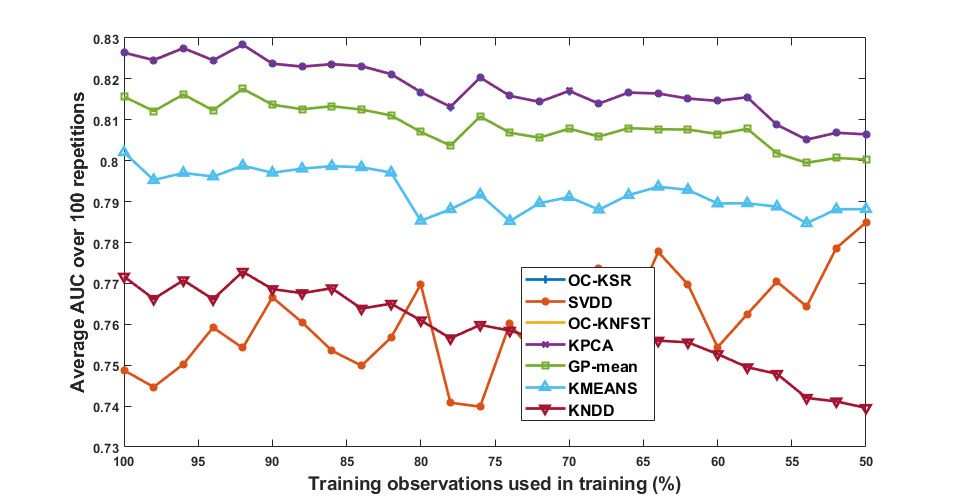}
\includegraphics[scale=.36]{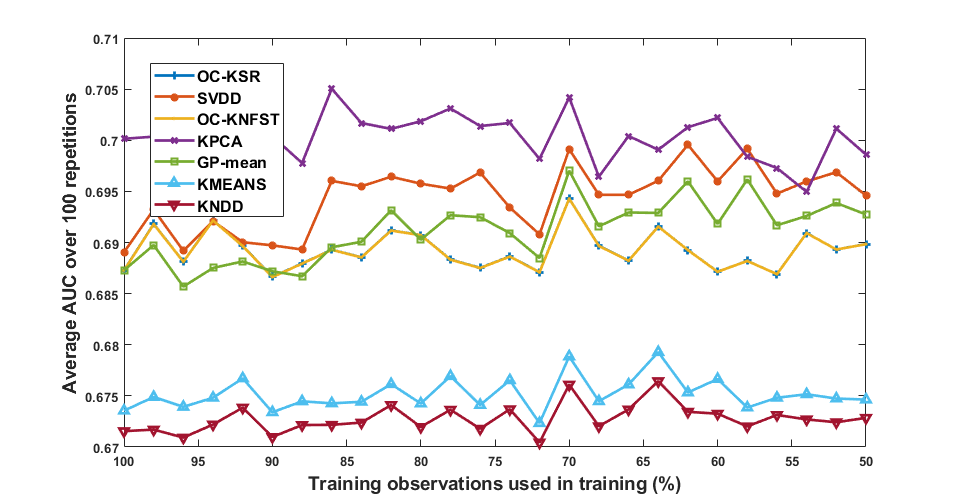}
\includegraphics[scale=.36]{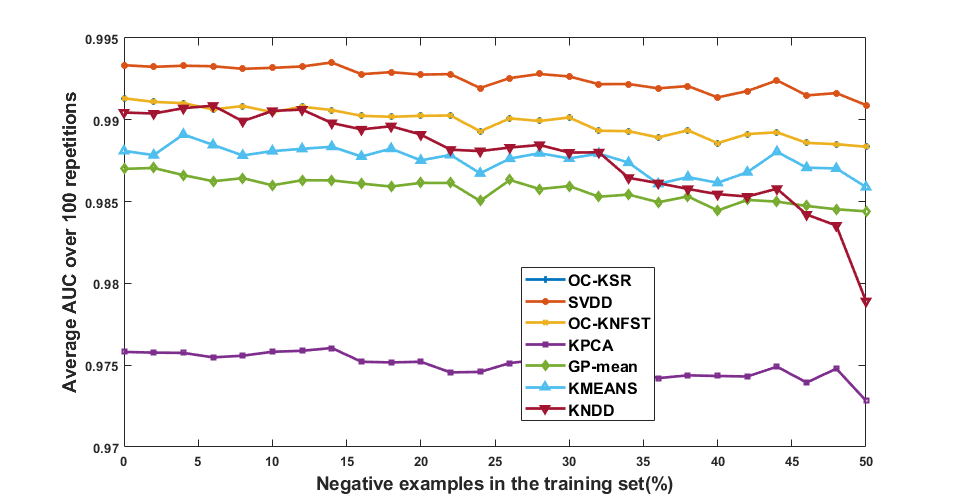}
\includegraphics[scale=.36]{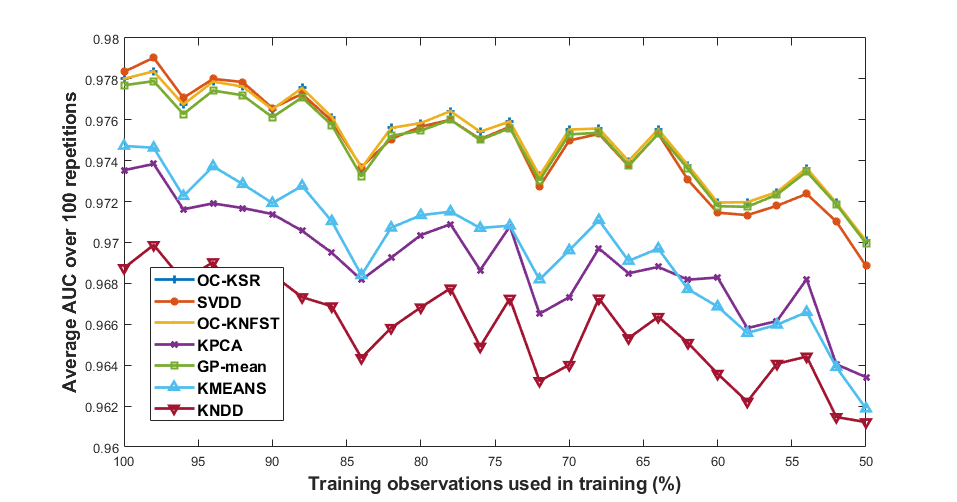}
\includegraphics[scale=.36]{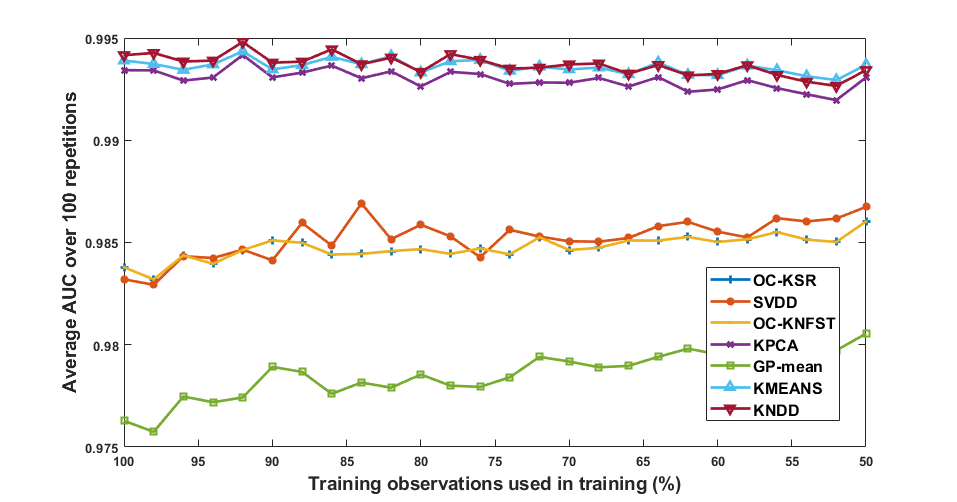}
\caption{The effect of training sample size on performance on datasets with medium to high dimensional feature vectors.(from top to bottom: Arcene, AD, Face, Caltech256 and MNIST)}
\label{r_fig1}
\end{figure}

\begin{figure}
\center
\includegraphics[scale=.36]{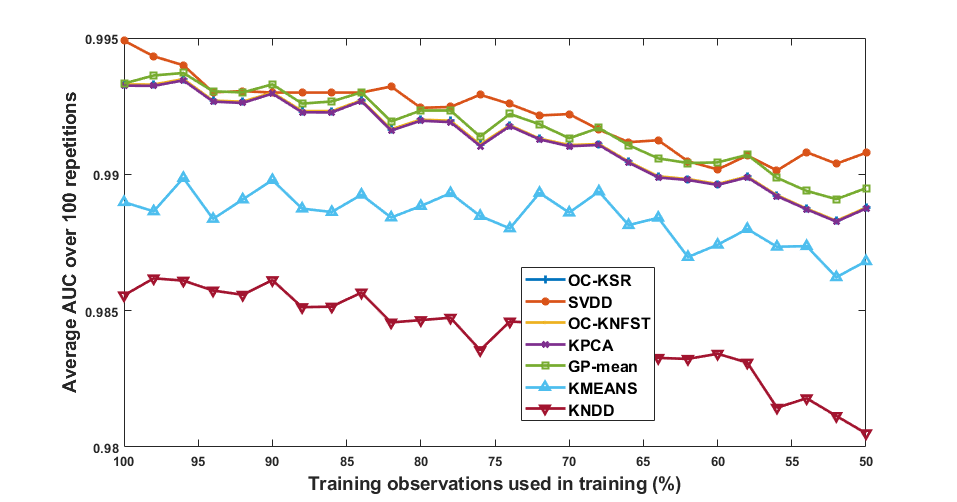}
\includegraphics[scale=.36]{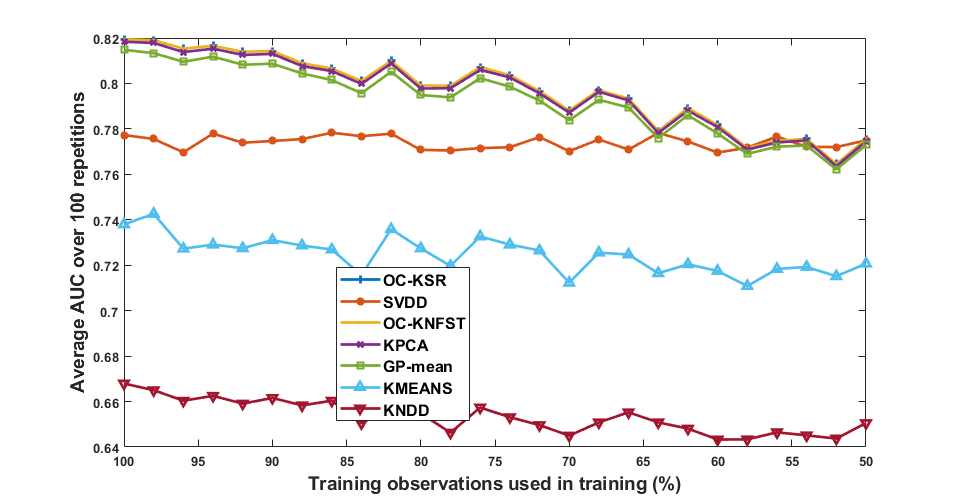}
\includegraphics[scale=.36]{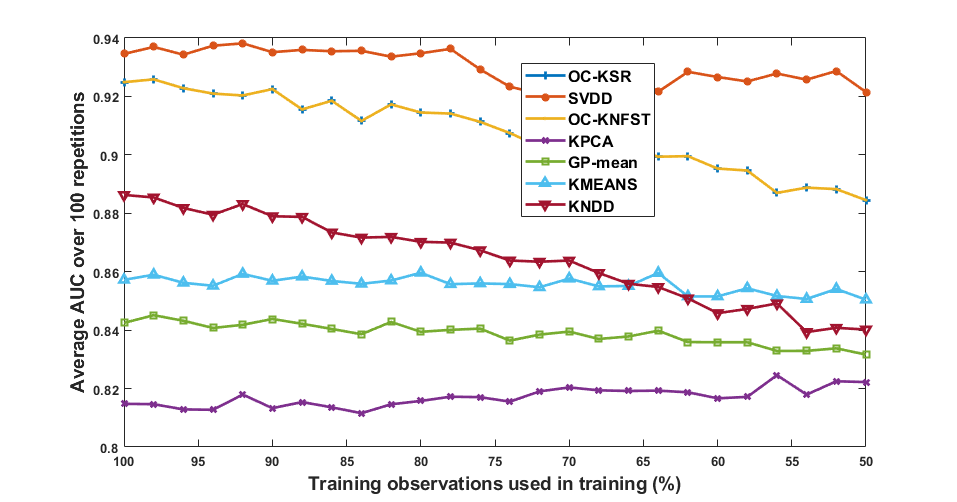}
\includegraphics[scale=.36]{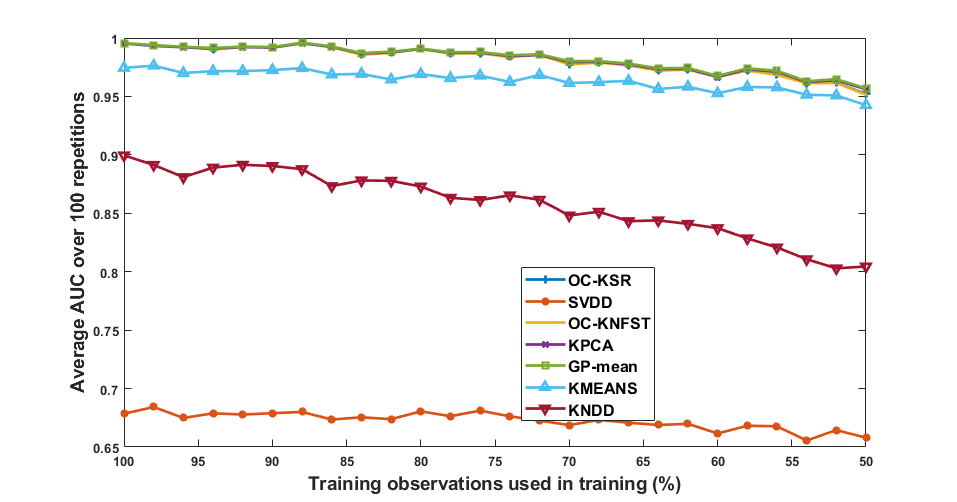}
\includegraphics[scale=.36]{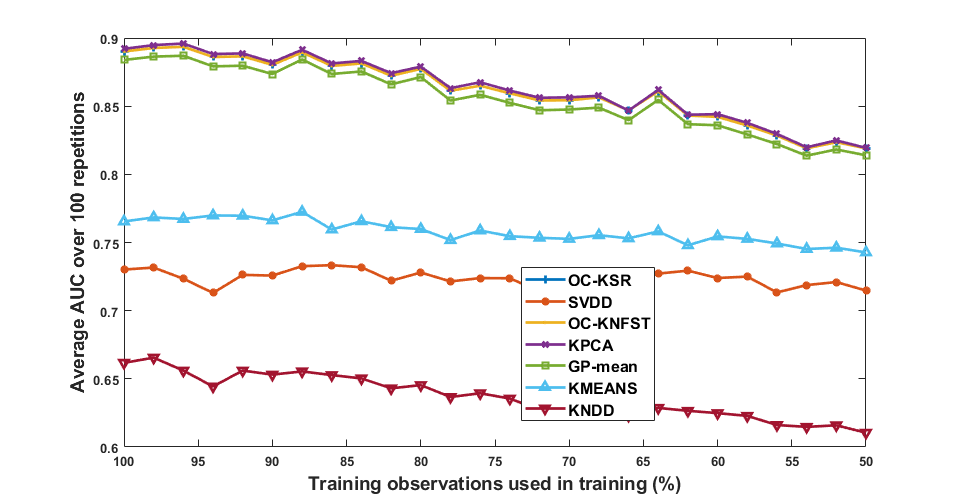}
\caption{The effect of training sample size on performance on datasets with relatively lower dimensional feature vectors.(from top to bottom: Pump, Sonar mines, Vehicle van, Vowel-0, Balance-scale)}
\label{r_fig2}
\end{figure}

In this experiment, the effect of training sample size on the performance of the proposed approach is compared to other methods. For this purpose, the training sample size is gradually decreased from $100\%$ of total training observations to $50\%$ in decrements of $2\%$. As the LOF method is found to perform much worse compared to others, it is excluded from this experiment. The results  are presented in Fig. \ref{r_fig1} and Fig \ref{r_fig2} for the medium to high and relatively lower dimensional datasets, respectively. As expected, for all the datasets, with the exception of the AD and MNIST, a reduction in training sample size deteriorates the performance of all systems. Regarding the AD dataset, an oscillatory behaviour is observed whereas for the MNIST dataset the performance of some methods (including the proposed OC-KSR approach) even slightly improves as the training set size decreases. This spurious behaviour will be the subject of future research.
%The behaviour might be attributed to the diversity of the training data the elimination of part of which %might lead to a more compact cluster in the observation space. It can be expected that a further %reduction in the training set for both the AD and MNIST datasets, would lead to a similar behaviour to that %of others.

Most importantly, the ranking of the proposed OC-KSR method in terms of average AUC as a function of training sample size is typically preserved. In particular, the proposed method achieves leading performance on 4 out the 10 datasets examined when using $100\%$ of the training data and continues to do so even when the training set is shrunk by up to $\approx 50\%$. A similar observation can be made regarding the datasets on which the proposed OC-KSR method ranked second, except for the AD dataset where an oscillation in performance is observed for the majority of the methods. It can be concluded that, although a reduction in the training sample size may degrade the performance of the proposed approach, it maintains its relative ranking position on the majority of the datasets.

\subsection{Using negative examples}
\begin{figure}
\center
\includegraphics[scale=.36]{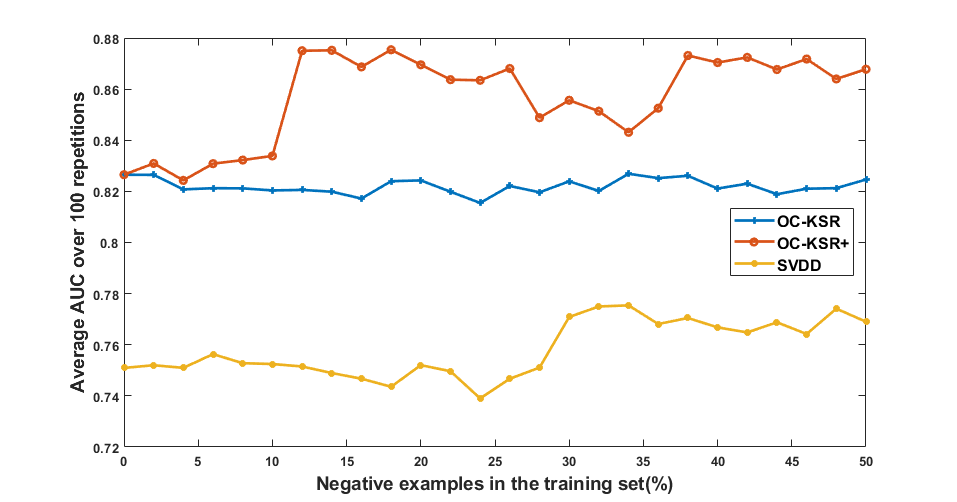}
\includegraphics[scale=.36]{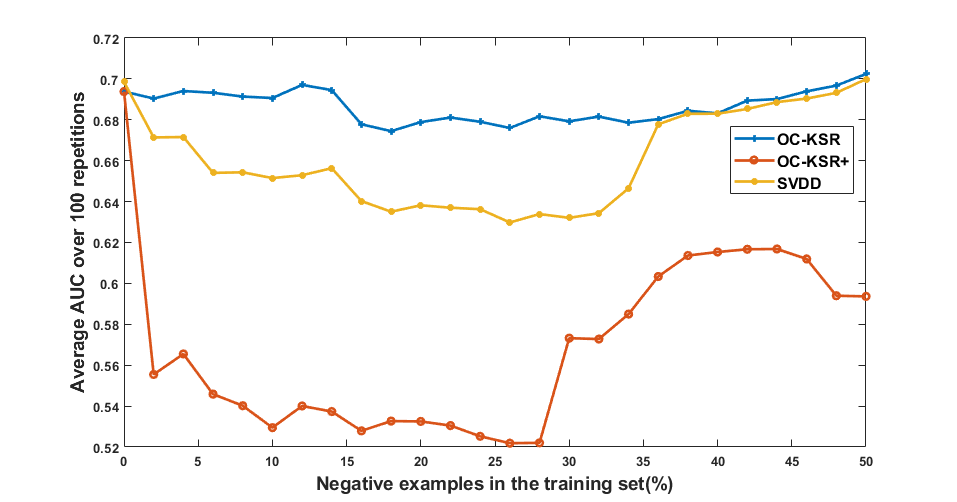}
\includegraphics[scale=.36]{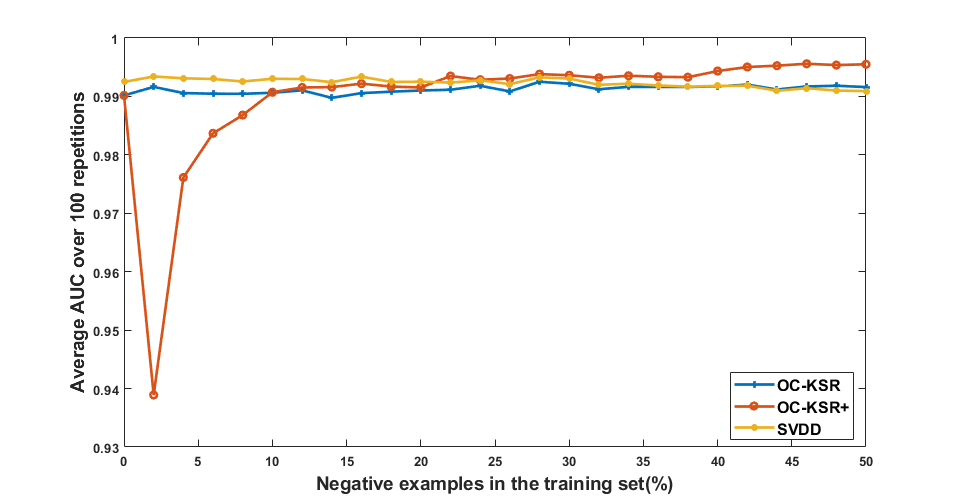}
\includegraphics[scale=.36]{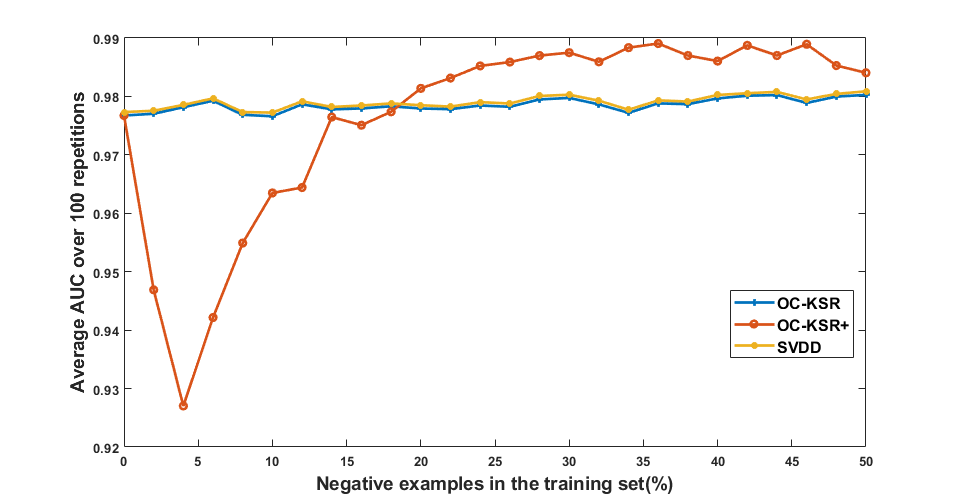}
\includegraphics[scale=.36]{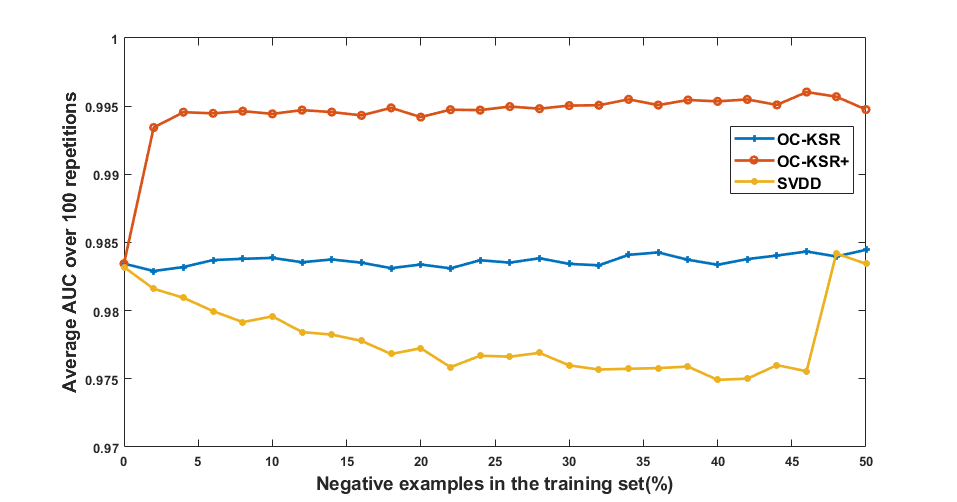}
\caption{The effect of using negative examples in the training set on datasets with medium to high dimensional feature vectors.(from top to bottom: Arcene, AD, Face, Caltech256 and MNIST)}
\label{cont_fig1}
\end{figure}

\begin{figure}
\center
\includegraphics[scale=.36]{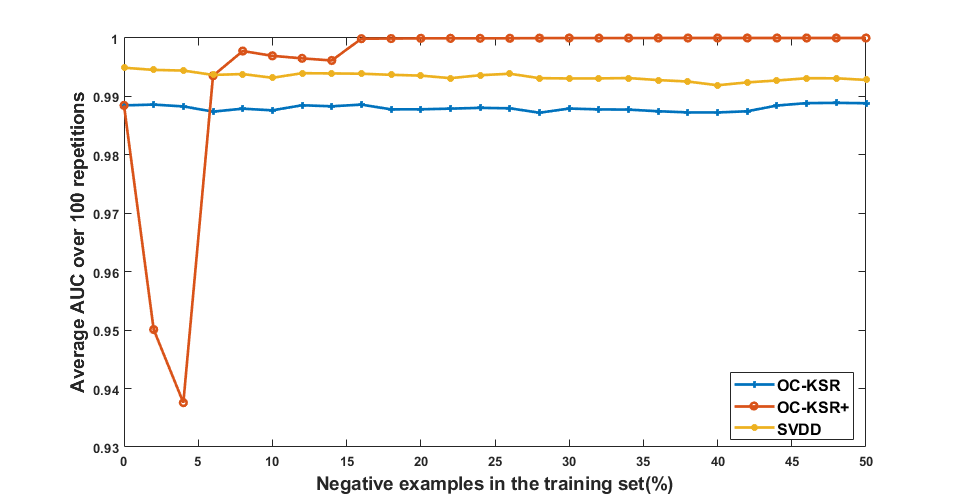}
\includegraphics[scale=.36]{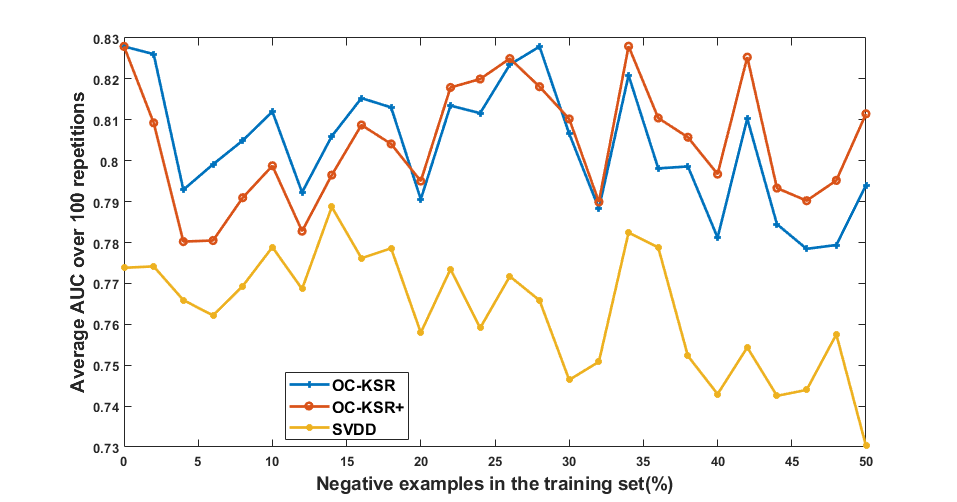}
\includegraphics[scale=.36]{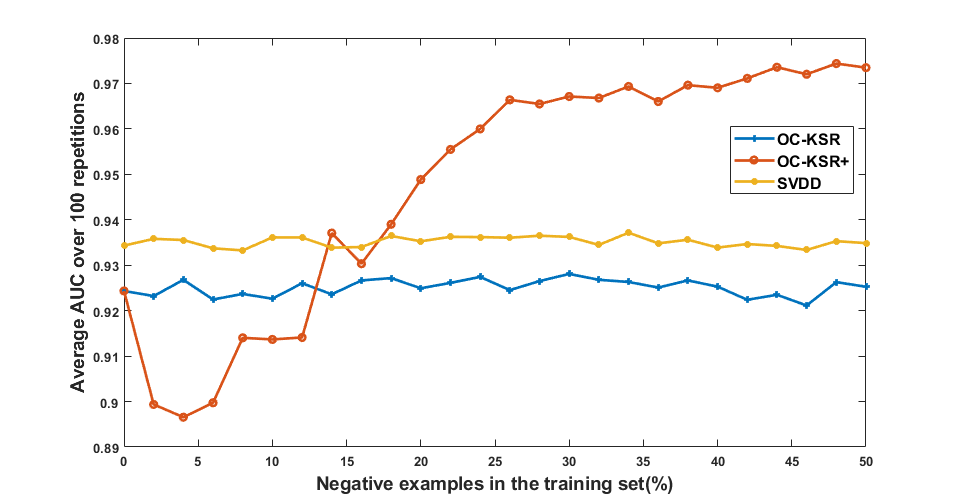}
\includegraphics[scale=.36]{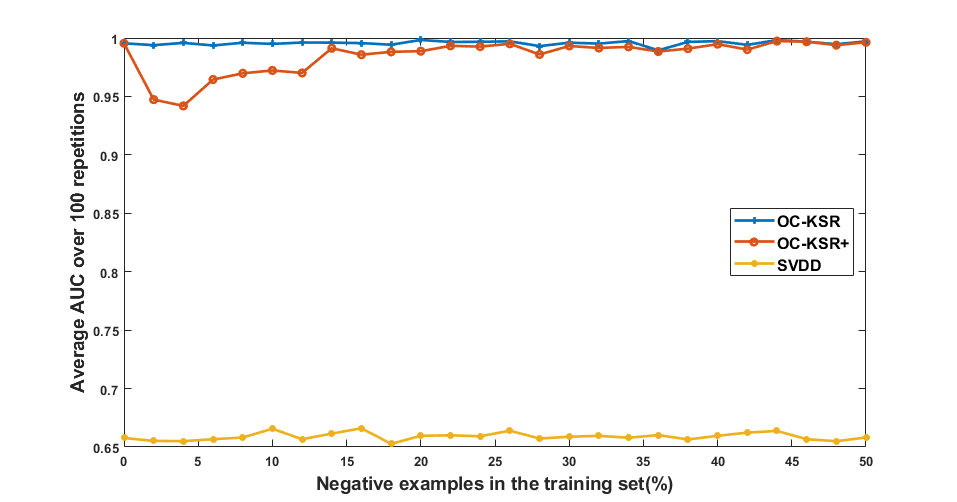}
\includegraphics[scale=.36]{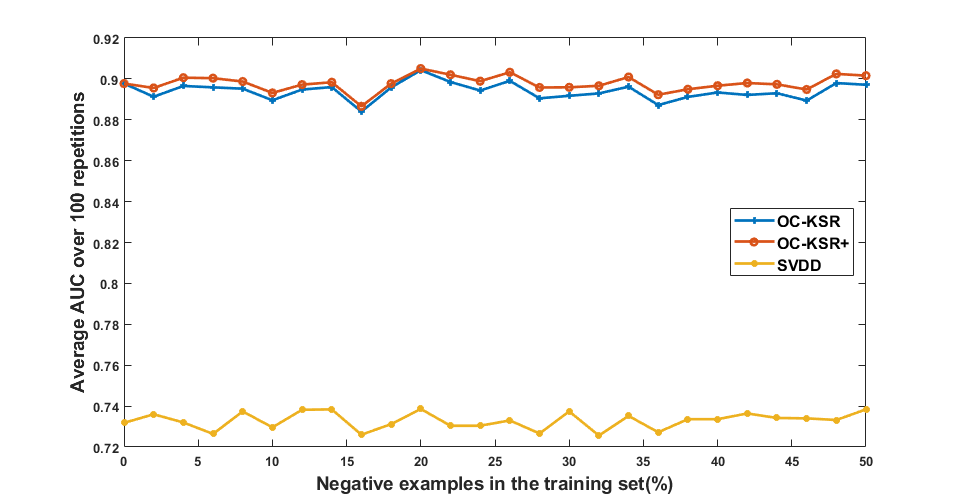}
\caption{The effect of using negative examples in the training set on datasets with relatively lower dimensional feature vectors.(from top to bottom: Pump, Sonar mines, Vehicle van, Vowel-0, Balance-scale)}
\label{cont_fig2}
\end{figure}

In this experiment, the effectiveness of the proposed approach in making use of counter-examples in the training set is examined. For this purpose, labelled negative samples are gradually included in the training set and the performances on different datasets are examined. On each dataset, the negative training examples are obtained from the negative samples of the corresponding dataset, the proportion of which relative to the initial positive sample set is increased from $0\%$ to $50\%$ in increments of $2\%$. Each experiment is repeated 100 times and the average AUC's are plotted. As among other methods only SVDD provides an explicit built-in mechanism for using negative examples in the training set, the methods included in this experiment are SVDD, the OC-KSR method without using negative examples (denoted as OC-KSR) and the OC-KSR method using counter-examples (denoted as OC-KSR$^+$). The results of this evaluation are depicted in Fig. \ref{cont_fig1} and Fig. \ref{cont_fig2} for the medium to high and relatively lower dimensional datasets, respectively. From the figures, the following observations can be made. Initially when negative examples are relatively much fewer than the positive samples, the OC-KSR$^+$ method does not seem to provide an advantage over OC-KSR. As more negative examples become available, the performance of the OC-KSR$^+$ method tends to improve. This is expected since when very few counter-examples are available (less than $10\%$ of the initial positive training set), the negative class may not be very well represented. Increasing the number of negative examples, they may better represent the non-target class and hence the OC-KSR$^+$ method outperforms OC-KSR in the majority of the datasets. Moreover, typically when more negative examples are available, the proposed OC-KSR$^+$ method also outperforms SVDD. The merits of the proposed OC-KSR$^+$ method over SVDD become more prominent as more and more negative examples are included in the training set.

\subsection{Computational complexity}
\label{comps}
In this section, the computational complexity of the proposed OC-KSR method in the training and test phases is discussed.
\subsubsection{Computational complexity in the training stage}
An analysis regarding the computational complexity of the proposed method in the training stage is as follows. As with all the kernel methods, the computation of the kernel matrix has a time complexity of $O(n^2d)$. Computing the additional part of the kernel matrix in the incremental scheme requires $O(dn\Delta n + d\Delta n^2)$ compound arithmetic operations each consisting of one addition and one multiplication (flam \cite{matrixdecom}), where $\Delta n$ is the number of additional training samples. The incremental Cholesky decomposition requires $\frac{1}{6}(n+\Delta n)^3-\frac{1}{6}n^3$. Given the Cholesky decomposition of $\mathbf{K}$, the linear equations $\mathbf{K}\boldsymbol{\alpha}^{opt}=\mathbf{y}^{opt}$ can be solved within $(n+\Delta n)^2$ flams. As a result, the computational cost of training the incremental OC-KSR approach in the updating phase is
\begin{IEEEeqnarray}{l}
\nonumber O(dn\Delta n + d\Delta n^2)+\frac{1}{6}(n+\Delta n)^3-\frac{1}{6}n^3+(n+\Delta n)^2\\
\nonumber = O(dn\Delta n + d\Delta n^2)+\frac{1}{2}n^2\Delta n+\frac{1}{2}n\Delta n^2+\frac{1}{6}\Delta n^3\\
\nonumber +(n+\Delta n)^2
\end{IEEEeqnarray}
\noindent assuming $\Delta n\ll n$, the cost can be approximated as
\begin{eqnarray}
(\frac{\Delta n}{2}+1)n^2
\end{eqnarray}
In the initial training stage, the computation of the kernel matrix, the Cholesky decomposition of the kernel matrix and solving $n$ linear equations are required. Noting that even in the initial stage the Cholesky decomposition can be performed in an incremental fashion (we assume $\Delta n=1$ during the initial training phase), the total cost in the initialisation stage can be approximated as
\begin{eqnarray}
O(n^2d)+\frac{3}{2}n^2
\end{eqnarray}
 As a result, if $d\gg \frac{3}{2}$ (which is often the case), the proposed algorithm would have a time complexity of $O(n^2d)$ in the training stage. That is, the computation of the kernel matrix has the dominant complexity in the training phase of the proposed approach.

\subsubsection{Computational complexity in the Test stage}
In the test phase, the OC-KSR method requires computation of $\mathbf{k}_z$ which has a time complexity of $O(nd)$ followed by the computation of $f(z)$ requiring $n$ flams. Hence the dominant computational complexity in the test phase is $O(nd)$. As the classification performance of the proposed approach is provably identical to the OC-KNFST method of \cite{6619277}, in the test phase the two methods are comparable.

Recently, an incremental variant of the OC-KNFST approach was proposed in \cite{8099922} which reduces the computational complexity of the original KNFST algorithm in the training stage.  Specifically, the incremental OC-KNFST algorithm requires $O(n^2d)+O(n^3)$ for the computation of the kernel matrix, its eigen-decomposition and matrix multiplications. As the computation of the kernel matrix is common for both the OC-KSR and the incremental OC-KNFST, the relative computational advantage of the OC-KSR over the incremental OC-KNFST in the training stage %when kernel matrix is pre-computed 
is $\approx \frac{2}{3}\frac{O(n^3)}{n^2}$. In other words, the computational superiority of the OC-KSR approach with respect to the incremental OC-KNFST increases almost linearly in the number of training samples, $n$. This is due to the fact that the method in \cite{8099922} uses eigen-decomposition, whereas OC-KSR solves the optimisation problem by regression.

The computational complexity of the method in \cite{8099922} in the updating phase of the training stage is $O(\Delta n^3+a n\Delta n ) \approx O(an\Delta n)$ where $a$ is the number of eigen-bases, upper bounded by $n$. As a result, in common scenarios where e.g. $\Delta n>10$, if the number of eigen-bases $a$ for the incremental OC-KNFST method exceeds $60\%$ of $n$, the proposed OC-KSR method would be more efficient. In summary, in the initial training phase, the proposed OC-KSR approach is computationally more efficient than the incremental OC-KNFST method of \cite{8099922}. In the updating phase, under mild conditions, it would be more efficient too.


\section{Conclusion}
\label{conclude}
A new nonlinear one-class classifier built upon the Fisher criterion was presented while providing a graph embedding view of the problem. The proposed OC-KSR approach operates by mapping the data onto a one-dimensional feature space where the scatter of training data is minimised while keeping positive samples far from the centre of the negative class. It was shown that positive and negative training observations were projected onto two distinct points in a feature subspace (the locations of which could be determined up to a multiplicative constant) yielding a kernel null-space Fisher analysis. The proposed method, unlike previous similar approaches, casts the problem under consideration into a regression framework optimising the criterion function via the efficient spectral regression method thus avoiding costly eigen-decomposition computations. It was illustrated that the dominant complexity of the proposed method in the training phase is the complexity of computing the kernel matrix. Moreover, the proposed OC-KSR approach offers a number of appealing characteristics such as the ability to be trained in an incremental fashion and the operability in an unsupervised mode. In addition, it was shown that in the presence of non-target training observations, such samples can be directly used to further refine the decision boundary for classification in a supervised mode. Extensive experiments conducted on several datasets with varied dimensions of features verified the merits of the proposed approach in comparison with some other alternatives.
% if have a single appendix:

% or
%\appendix  % for no appendix heading
% do not use \section anymore after \appendix, only \section*
% is possibly needed

% use appendices with more than one appendix
% then use \section to start each appendix
% you must declare a \section before using any
% \subsection or using \label (\appendices by itself
% starts a section numbered zero.)
%

% you can choose not to have a title for an appendix
% if you want by leaving the argument blank

% use section* for acknowledgement
\section*{Acknowledgment}

% Can use something like this to put references on a page
% by themselves when using endfloat and the captionsoff option.
\ifCLASSOPTIONcaptionsoff
  \newpage
\fi

% trigger a \newpage just before the given reference
% number - used to balance the columns on the last page
% adjust value as needed - may need to be readjusted if
% the document is modified later
%\IEEEtriggeratref{8}
% The "triggered" command can be changed if desired:
%\IEEEtriggercmd{\enlargethispage{-5in}}

% references section

% can use a bibliography generated by BibTeX as a .bbl file
% BibTeX documentation can be easily obtained at:
% http://www.ctan.org/tex-archive/biblio/bibtex/contrib/doc/
% The IEEEtran BibTeX style support page is at:
% http://www.michaelshell.org/tex/ieeetran/bibtex/
%\bibliographystyle{IEEEtran}
% argument is your BibTeX string definitions and bibliography database(s)
%\bibliography{IEEEabrv,../bib/paper}
%
% <OR> manually copy in the resultant .bbl file
% set second argument of \begin to the number of references
% (used to reserve space for the reference number labels box)
\bibliographystyle{IEEEtran}
\bibliography{IEEEexample2}

% Generated by IEEEtran.bst, version: 1.14 (2015/08/26)
\begin{thebibliography}{10}
\providecommand{\url}[1]{#1}
\csname url@samestyle\endcsname
\providecommand{\newblock}{\relax}
\providecommand{\bibinfo}[2]{#2}
\providecommand{\BIBentrySTDinterwordspacing}{\spaceskip=0pt\relax}
\providecommand{\BIBentryALTinterwordstretchfactor}{4}
\providecommand{\BIBentryALTinterwordspacing}{\spaceskip=\fontdimen2\font plus
\BIBentryALTinterwordstretchfactor\fontdimen3\font minus
  \fontdimen4\font\relax}
\providecommand{\BIBforeignlanguage}[2]{{%
\expandafter\ifx\csname l@#1\endcsname\relax
\typeout{** WARNING: IEEEtran.bst: No hyphenation pattern has been}%
\typeout{** loaded for the language `#1'. Using the pattern for}%
\typeout{** the default language instead.}%
\else
\language=\csname l@#1\endcsname
\fi
#2}}
\providecommand{\BIBdecl}{\relax}
\BIBdecl

\bibitem{khan_madden_2014}
S.~S. Khan and M.~G. Madden, ``One-class classification: taxonomy of study and
  review of techniques,'' \emph{The Knowledge Engineering Review}, vol.~29,
  no.~3, pp. 345--374, 2014.

\bibitem{6846360}
P.~Nader, P.~Honeine, and P.~Beauseroy, ``${l_p}$-norms in one-class
  classification for intrusion detection in scada systems,'' \emph{IEEE
  Transactions on Industrial Informatics}, vol.~10, no.~4, pp. 2308--2317, Nov
  2014.

\bibitem{BEGHI20141953}
A.~Beghi, L.~Cecchinato, C.~Corazzol, M.~Rampazzo, F.~Simmini, and G.~Susto,
  ``A one-class svm based tool for machine learning novelty detection in hvac
  chiller systems,'' \emph{IFAC Proceedings Volumes}, vol.~47, no.~3, pp. 1953
  -- 1958, 2014, 19th IFAC World Congress.

\bibitem{4694106}
S.~Budalakoti, A.~N. Srivastava, and M.~E. Otey, ``Anomaly detection and
  diagnosis algorithms for discrete symbol sequences with applications to
  airline safety,'' \emph{IEEE Transactions on Systems, Man, and Cybernetics,
  Part C (Applications and Reviews)}, vol.~39, no.~1, pp. 101--113, Jan 2009.

\bibitem{Kamaruddin:2016:CCF:2980258.2980319}
S.~Kamaruddin and V.~Ravi, ``Credit card fraud detection using big data
  analytics: Use of psoaann based one-class classification,'' in
  \emph{Proceedings of the International Conference on Informatics and
  Analytics}, ser. ICIA-16.\hskip 1em plus 0.5em minus 0.4em\relax New York,
  NY, USA: ACM, 2016, pp. 33:1--33:8.

\bibitem{7435726}
G.~G. Sundarkumar, V.~Ravi, and V.~Siddeshwar, ``One-class support vector
  machine based undersampling: Application to churn prediction and insurance
  fraud detection,'' in \emph{2015 IEEE International Conference on
  Computational Intelligence and Computing Research (ICCIC)}, Dec 2015, pp.
  1--7.

\bibitem{6566012}
M.~Yu, Y.~Yu, A.~Rhuma, S.~M.~R. Naqvi, L.~Wang, and J.~A. Chambers, ``An
  online one class support vector machine-based person-specific fall detection
  system for monitoring an elderly individual in a room environment,''
  \emph{IEEE Journal of Biomedical and Health Informatics}, vol.~17, no.~6, pp.
  1002--1014, Nov 2013.

\bibitem{4668357}
A.~Rabaoui, M.~Davy, S.~Rossignol, and N.~Ellouze, ``Using one-class svms and
  wavelets for audio surveillance,'' \emph{IEEE Transactions on Information
  Forensics and Security}, vol.~3, no.~4, pp. 763--775, Dec 2008.

\bibitem{8386786}
V.~L. Cao, M.~Nicolau, and J.~McDermott, ``Learning neural representations for
  network anomaly detection,'' \emph{IEEE Transactions on Cybernetics}, pp.
  1--14, 2018.

\bibitem{Minter1975}
T.~Minter, ``Single-class classification,'' in \emph{Symposium on Machine
  Processing of Remotely Sensed Data}.\hskip 1em plus 0.5em minus 0.4em\relax
  Indiana, USA: IEEE, 1975, pp. 2A12--2A15.

\bibitem{1993STIN9324043M}
M.~M. {Moya}, M.~W. {Koch}, and L.~D. {Hostetler}, ``{One-class classifier
  networks for target recognition applications},'' \emph{NASA STI/Recon
  Technical Report N}, vol.~93, 1993.

\bibitem{RITTER1997525}
G.~Ritter and M.~T. Gallegos, ``Outliers in statistical pattern recognition and
  an application to automatic chromosome classification,'' \emph{Pattern
  Recognition Letters}, vol.~18, no.~6, pp. 525 -- 539, 1997.

\bibitem{318023}
C.~M. Bishop, ``Novelty detection and neural network validation,'' \emph{IEE
  Proceedings - Vision, Image and Signal Processing}, vol. 141, no.~4, pp.
  217--222, Aug 1994.

\bibitem{Japkowicz:1999:CLA:929980}
N.~Japkowicz, ``Concept learning in the absence of counterexamples: An
  autoassociation-based approach to classification,'' Ph.D. dissertation, New
  Brunswick, NJ, USA, 1999, aAI9947599.

\bibitem{6809169}
W.~J. Scheirer, L.~P. Jain, and T.~E. Boult, ``Probability models for open set
  recognition,'' \emph{IEEE Transactions on Pattern Analysis and Machine
  Intelligence}, vol.~36, no.~11, pp. 2317--2324, Nov 2014.

\bibitem{30e27ea86}
D.~Tax, ``One-class classification; concept-learning in the absence of
  counter-examples,'' Ph.D. dissertation, Delft University of Technology, 2001,
  aSCI Dissertation Series 65.

\bibitem{PIMENTEL2014215}
M.~A. Pimentel, D.~A. Clifton, L.~Clifton, and L.~Tarassenko, ``A review of
  novelty detection,'' \emph{Signal Processing}, vol.~99, pp. 215 -- 249, 2014.

\bibitem{6636290}
J.~Kittler, W.~Christmas, T.~de~Campos, D.~Windridge, F.~Yan, J.~Illingworth,
  and M.~Osman, ``Domain anomaly detection in machine perception: A system
  architecture and taxonomy,'' \emph{IEEE Transactions on Pattern Analysis and
  Machine Intelligence}, vol.~36, no.~5, pp. 845--859, May 2014.

\bibitem{Tax:2001:COC:648055.744087}
D.~M.~J. Tax and R.~P.~W. Duin, ``Combining one-class classifiers,'' in
  \emph{Proceedings of the Second International Workshop on Multiple Classifier
  Systems}, ser. MCS '01.\hskip 1em plus 0.5em minus 0.4em\relax London, UK,
  UK: Springer-Verlag, 2001, pp. 299--308.

\bibitem{10.1137/1.9781611973440.67}
\BIBentryALTinterwordspacing
L.~Friedland, A.~Gentzel, and D.~Jensen, \emph{Classifier-Adjusted Density
  Estimation for Anomaly Detection and One-Class Classification}, pp. 578--586.
  [Online]. Available:
  \url{https://epubs.siam.org/doi/abs/10.1137/1.9781611973440.67}
\BIBentrySTDinterwordspacing

\bibitem{HOFFMANN2007863}
H.~Hoffmann, ``Kernel pca for novelty detection,'' \emph{Pattern Recognition},
  vol.~40, no.~3, pp. 863 -- 874, 2007.

\bibitem{7492368}
M.~Sabokrou, M.~Fathy, and M.~Hoseini, ``Video anomaly detection and
  localisation based on the sparsity and reconstruction error of
  auto-encoder,'' \emph{Electronics Letters}, vol.~52, no.~13, pp. 1122--1124,
  2016.

\bibitem{7984788}
S.~R. Arashloo, J.~Kittler, and W.~Christmas, ``An anomaly detection approach
  to face spoofing detection: A new formulation and evaluation protocol,''
  \emph{IEEE Access}, vol.~5, pp. 13\,868--13\,882, 2017.

\bibitem{7393467}
B.~Song, P.~Li, J.~Li, and A.~Plaza, ``One-class classification of remote
  sensing images using kernel sparse representation,'' \emph{IEEE Journal of
  Selected Topics in Applied Earth Observations and Remote Sensing}, vol.~9,
  no.~4, pp. 1613--1623, April 2016.

\bibitem{Tax2004}
\BIBentryALTinterwordspacing
D.~M. Tax and R.~P. Duin, ``Support vector data description,'' \emph{Machine
  Learning}, vol.~54, no.~1, pp. 45--66, Jan 2004. [Online]. Available:
  \url{https://doi.org/10.1023/B:MACH.0000008084.60811.49}
\BIBentrySTDinterwordspacing

\bibitem{Scholkopf:2001:ESH:1119748.1119749}
\BIBentryALTinterwordspacing
B.~Sch\"{o}lkopf, J.~C. Platt, J.~C. Shawe-Taylor, A.~J. Smola, and R.~C.
  Williamson, ``Estimating the support of a high-dimensional distribution,''
  \emph{Neural Comput.}, vol.~13, no.~7, pp. 1443--1471, Jul. 2001. [Online].
  Available: \url{https://doi.org/10.1162/089976601750264965}
\BIBentrySTDinterwordspacing

\bibitem{6869001}
Y.~Xiao, H.~Wang, and W.~Xu, ``Parameter selection of gaussian kernel for
  one-class svm,'' \emph{IEEE Transactions on Cybernetics}, vol.~45, no.~5, pp.
  941--953, May 2015.

\bibitem{efddef0571dc4c9594328f21683c3d45}
E.~Pekalska, D.~Tax, R.~Duin, S.~Becker, S.~Thrun, and K.~Obermayer,
  \emph{One-Class LP Classifiers for Dissimilarity Representations}.\hskip 1em
  plus 0.5em minus 0.4em\relax United States: MIT Press, 2002, pp. 761--768.

\bibitem{10.1007/978-3-642-21557-5_13}
P.~Casale, O.~Pujol, and P.~Radeva, ``Approximate convex hulls family for
  one-class classification,'' in \emph{Multiple Classifier Systems},
  C.~Sansone, J.~Kittler, and F.~Roli, Eds.\hskip 1em plus 0.5em minus
  0.4em\relax Berlin, Heidelberg: Springer Berlin Heidelberg, 2011, pp.
  106--115.

\bibitem{8125573}
D.~Fernández-Francos, .~Fontenla-Romero, and A.~Alonso-Betanzos, ``One-class
  convex hull-based algorithm for classification in distributed environments,''
  \emph{IEEE Transactions on Systems, Man, and Cybernetics: Systems}, pp.
  1--11, 2017.

\bibitem{10.1007/978-1-4471-1599-1_110}
A.~Ypma and R.~P.~W. Duin, ``Support objects for domain approximation,'' in
  \emph{ICANN 98}, L.~Niklasson, M.~Bod{\'e}n, and T.~Ziemke, Eds.\hskip 1em
  plus 0.5em minus 0.4em\relax London: Springer London, 1998, pp. 719--724.

\bibitem{DBLP:journals/corr/abs-1802-09088}
\BIBentryALTinterwordspacing
M.~Sabokrou, M.~Khalooei, M.~Fathy, and E.~Adeli, ``Adversarially learned
  one-class classifier for novelty detection,'' \emph{CoRR}, vol.
  abs/1802.09088, 2018. [Online]. Available:
  \url{http://arxiv.org/abs/1802.09088}
\BIBentrySTDinterwordspacing

\bibitem{DBLP:journals/corr/abs-1801-05365}
\BIBentryALTinterwordspacing
P.~Perera and V.~M. Patel, ``Learning deep features for one-class
  classification,'' \emph{CoRR}, vol. abs/1801.05365, 2018. [Online].
  Available: \url{http://arxiv.org/abs/1801.05365}
\BIBentrySTDinterwordspacing

\bibitem{7938706}
S.~Wang, Q.~Liu, E.~Zhu, J.~Yin, and W.~Zhao, ``Mst-gen: An efficient parameter
  selection method for one-class extreme learning machine,'' \emph{IEEE
  Transactions on Cybernetics}, vol.~47, no.~10, pp. 3266--3279, Oct 2017.

\bibitem{8293843}
S.~S. Khan and A.~Ahmad, ``Relationship between variants of one-class nearest
  neighbors and creating their accurate ensembles,'' \emph{IEEE Transactions on
  Knowledge and Data Engineering}, vol.~30, no.~9, pp. 1796--1809, Sep. 2018.

\bibitem{NIPS2004_2656}
V.~Roth, ``Outlier detection with one-class kernel fisher discriminants,'' in
  \emph{Advances in Neural Information Processing Systems 17}, L.~K. Saul,
  Y.~Weiss, and L.~Bottou, Eds.\hskip 1em plus 0.5em minus 0.4em\relax MIT
  Press, 2005, pp. 1169--1176.

\bibitem{Roth:2006:KFD:1117520.1117529}
------, ``Kernel fisher discriminants for outlier detection,'' \emph{Neural
  Comput.}, vol.~18, no.~4, pp. 942--960, Apr. 2006.

\bibitem{6619277}
P.~Bodesheim, A.~Freytag, E.~Rodner, M.~Kemmler, and J.~Denzler, ``Kernel null
  space methods for novelty detection,'' in \emph{2013 IEEE Conference on
  Computer Vision and Pattern Recognition}, June 2013, pp. 3374--3381.

\bibitem{6857384}
F.~Dufrenois, ``A one-class kernel fisher criterion for outlier detection,''
  \emph{IEEE Transactions on Neural Networks and Learning Systems}, vol.~26,
  no.~5, pp. 982--994, May 2015.

\bibitem{6384805}
F.~Dufrenois and J.~C. Noyer, ``Formulating robust linear regression estimation
  as a one-class lda criterion: Discriminative hat matrix,'' \emph{IEEE
  Transactions on Neural Networks and Learning Systems}, vol.~24, no.~2, pp.
  262--273, Feb 2013.

\bibitem{kernelfisher}
S.~Mika, G.~Ratsch, J.~Weston, B.~Scholkopf, and K.~R. Mullers, ``{Fisher
  discriminant analysis with kernels},'' \emph{Neural Networks for Signal
  Processing IX, 1999. Proceedings of the 1999 IEEE Signal Processing Society
  Workshop}, pp. 41--48, Aug. 1999.

\bibitem{Gkernel}
G.~Baudat and F.~Anouar, ``Generalized discriminant analysis using a kernel
  approach.'' \emph{Neural Computation}, vol.~12, no.~10, pp. 2385--2404, 2000.

\bibitem{kernelbook}
B.~Scholkopf and A.~J. Smola, \emph{Learning with Kernels: Support Vector
  Machines, Regularization, Optimization, and Beyond}.\hskip 1em plus 0.5em
  minus 0.4em\relax Cambridge, MA, USA: MIT Press, 2001.

\bibitem{cortes1995support}
C.~Cortes and V.~Vapnik, ``Support vector networks,'' \emph{Machine Learning},
  vol.~20, pp. 273--297, 1995.

\bibitem{7727608}
F.~Dufrenois and J.~C. Noyer, ``A null space based one class kernel fisher
  discriminant,'' in \emph{2016 International Joint Conference on Neural
  Networks (IJCNN)}, July 2016, pp. 3203--3210.

\bibitem{8099922}
J.~Liu, Z.~Lian, Y.~Wang, and J.~Xiao, ``Incremental kernel null space
  discriminant analysis for novelty detection,'' in \emph{2017 IEEE Conference
  on Computer Vision and Pattern Recognition (CVPR)}, July 2017, pp.
  4123--4131.

\bibitem{SRKDA}
D.~Cai, X.~He, and J.~Han, ``Speed up kernel discriminant analysis,'' \emph{The
  VLDB Journal}, vol.~20, no.~1, pp. 21--33, Feb. 2011.

\bibitem{GueSchVis07}
S.~G\"unter, N.~N. Schraudolph, and S.~V.~N. Vishwanathan, ``Fast iterative
  kernel principal component analysis,'' vol.~8, pp. 1893--1918, 2007.

\bibitem{4522554}
N.~Kwak, ``Principal component analysis based on l1-norm maximization,''
  \emph{IEEE Transactions on Pattern Analysis and Machine Intelligence},
  vol.~30, no.~9, pp. 1672--1680, Sept 2008.

\bibitem{6795824}
V.~Roth, ``Kernel fisher discriminants for outlier detection,'' \emph{Neural
  Computation}, vol.~18, no.~4, pp. 942--960, April 2006.

\bibitem{7045967}
P.~Bodesheim, A.~Freytag, E.~Rodner, and J.~Denzler, ``Local novelty detection
  in multi-class recognition problems,'' in \emph{2015 IEEE Winter Conference
  on Applications of Computer Vision}, Jan 2015, pp. 813--820.

\bibitem{DUFRENOIS201696}
F.~Dufrenois and J.~Noyer, ``One class proximal support vector machines,''
  \emph{Pattern Recognition}, vol.~52, pp. 96 -- 112, 2016.

\bibitem{6905848}
S.~R. Arashloo and J.~Kittler, ``Class-specific kernel fusion of multiple
  descriptors for face verification using multiscale binarised statistical
  image features,'' \emph{IEEE Transactions on Information Forensics and
  Security}, vol.~9, no.~12, pp. 2100--2109, Dec 2014.

\bibitem{10.1007/3-540-44581}
B.~Sch{\"o}lkopf, R.~Herbrich, and A.~J. Smola, ``A generalized representer
  theorem,'' in \emph{Computational Learning Theory}, D.~Helmbold and
  B.~Williamson, Eds.\hskip 1em plus 0.5em minus 0.4em\relax Berlin,
  Heidelberg: Springer Berlin Heidelberg, 2001, pp. 416--426.

\bibitem{8294302}
H.~Cai, V.~W. Zheng, and K.~Chang, ``A comprehensive survey of graph embedding:
  Problems, techniques and applications,'' \emph{IEEE Transactions on Knowledge
  and Data Engineering}, pp. 1--1, 2018.

\bibitem{8047276}
Q.~Wang, Z.~Mao, B.~Wang, and L.~Guo, ``Knowledge graph embedding: A survey of
  approaches and applications,'' \emph{IEEE Transactions on Knowledge and Data
  Engineering}, vol.~29, no.~12, pp. 2724--2743, Dec 2017.

\bibitem{matrixdecom}
{G.W. Stewart}, \emph{Matrix algorithms -- Volume I: Basic
  decompositions}.\hskip 1em plus 0.5em minus 0.4em\relax SIAM, 2001.

\bibitem{UCIwebsite}
\BIBentryALTinterwordspacing
{UCI Machine Learning Repository}. [Online]. Available:
  \url{https://archive.ics.uci.edu/ml/datasets.html}
\BIBentrySTDinterwordspacing

\bibitem{8096133}
S.~Tirunagari, S.~Kouchaki, D.~Abasolo, and N.~Poh, ``One dimensional local
  binary patterns of electroencephalogram signals for detecting alzheimer's
  disease,'' in \emph{2017 22nd International Conference on Digital Signal
  Processing (DSP)}, Aug 2017, pp. 1--5.

\bibitem{7298594}
C.~Szegedy, W.~Liu, Y.~Jia, P.~Sermanet, S.~Reed, D.~Anguelov, D.~Erhan,
  V.~Vanhoucke, and A.~Rabinovich, ``Going deeper with convolutions,'' in
  \emph{2015 IEEE Conference on Computer Vision and Pattern Recognition
  (CVPR)}, June 2015, pp. 1--9.

\bibitem{Costa-Pazo_BIOSIG2016_2016}
A.~Costa-Pazo, S.~Bhattacharjee, E.~Vazquez-Fernandez, and S.~Marcel, ``The
  replay-mobile face presentation-attack database,'' in \emph{Proceedings of
  the International Conference on Biometrics Special Interests Group (BioSIG)},
  Sep. 2016.

\bibitem{griffinHolubPerona}
\BIBentryALTinterwordspacing
G.~Griffin, A.~Holub, and P.~Perona, ``Caltech-256 object category dataset,''
  California Institute of Technology, Tech. Rep. 7694, 2007. [Online].
  Available: \url{http://authors.library.caltech.edu/7694}
\BIBentrySTDinterwordspacing

\bibitem{726791}
Y.~Lecun, L.~Bottou, Y.~Bengio, and P.~Haffner, ``Gradient-based learning
  applied to document recognition,'' \emph{Proceedings of the IEEE}, vol.~86,
  no.~11, pp. 2278--2324, Nov 1998.

\bibitem{delftwebsite}
\BIBentryALTinterwordspacing
{Delft University Archive of One-Class Data Sets}. [Online]. Available:
  \url{http://homepage.tudelft.nl/n9d04/occ/index.html}
\BIBentrySTDinterwordspacing

\bibitem{KEMMLER20133507}
M.~Kemmler, E.~Rodner, E.-S. Wacker, and J.~Denzler, ``One-class classification
  with gaussian processes,'' \emph{Pattern Recognition}, vol.~46, no.~12, pp.
  3507 -- 3518, 2013.

\bibitem{Breunig00lof:identifying}
M.~Breunig, H.-P. Kriegel, R.~T. Ng, and J.~Sander, ``Lof: Identifying
  density-based local outliers,'' in \emph{PROCEEDINGS OF THE 2000 ACM SIGMOD
  INTERNATIONAL CONFERENCE ON MANAGEMENT OF DATA}.\hskip 1em plus 0.5em minus
  0.4em\relax ACM, 2000, pp. 93--104.

\bibitem{DERRAC20113}
J.~Derrac, S.~García, D.~Molina, and F.~Herrera, ``A practical tutorial on the
  use of nonparametric statistical tests as a methodology for comparing
  evolutionary and swarm intelligence algorithms,'' \emph{Swarm and
  Evolutionary Computation}, vol.~1, no.~1, pp. 3 -- 18, 2011.

\end{thebibliography}

% biography section
% 
% If you have an EPS/PDF photo (graphicx package needed) extra braces are
% needed around the contents of the optional argument to biography to prevent
% the LaTeX parser from getting confused when it sees the complicated
% \includegraphics command within an optional argument. (You could create
% your own custom macro containing the \includegraphics command to make things
% simpler here.)
% or if you just want to reserve a space for a photo:

% if you will not have a photo at all:

% insert where needed to balance the two columns on the last page with
% biographies
%\newpage

% You can push biographies down or up by placing
% a \vfill before or after them. The appropriate
% use of \vfill depends on what kind of text is
% on the last page and whether or not the columns
% are being equalized.

%\vfill

% Can be used to pull up biographies so that the bottom of the last one
% is flush with the other column.
%\enlargethispage{-5in}

\begin{IEEEbiography}[{\includegraphics[scale=.6]{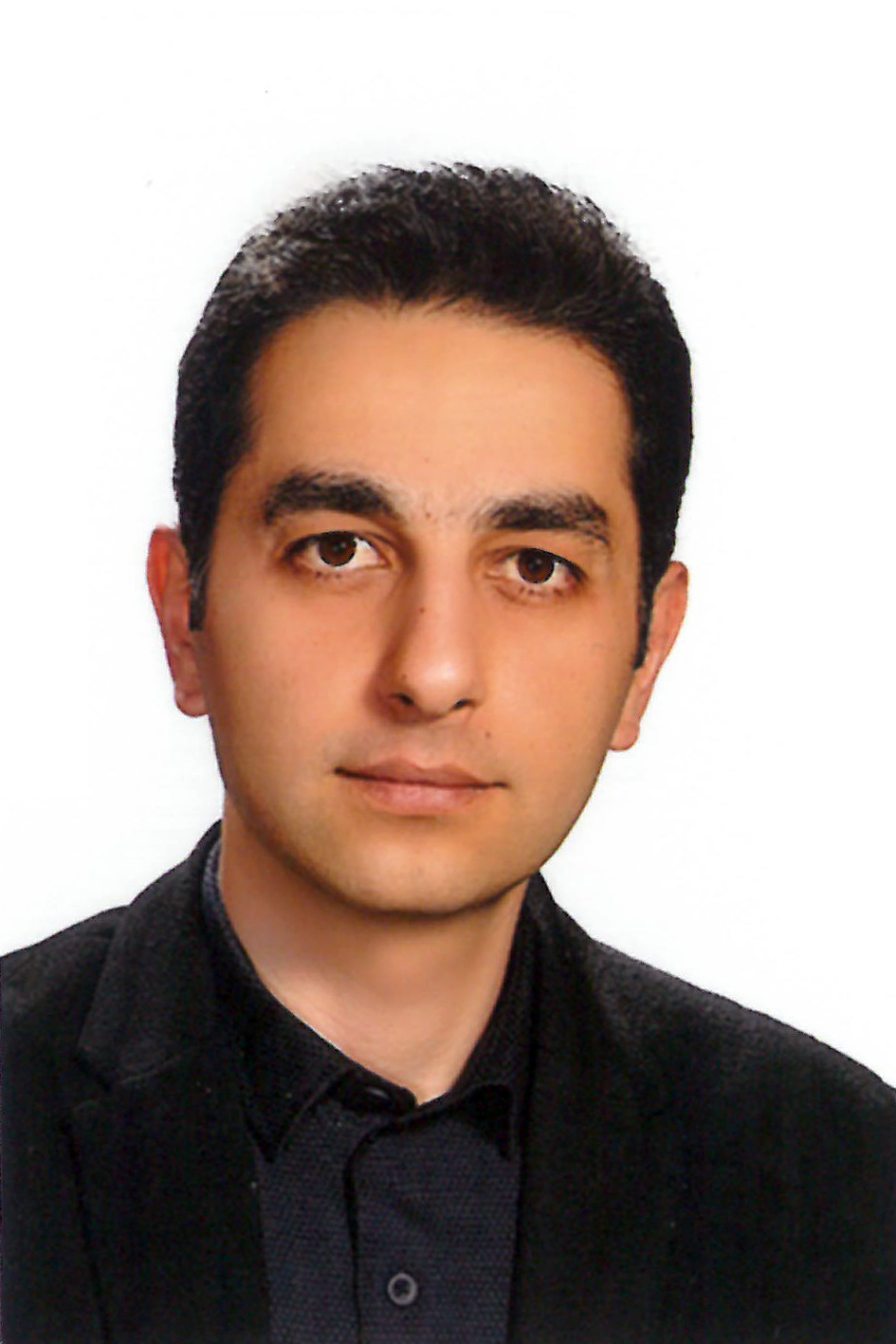}}]{Shervin Rahimzadeh Arashloo}
received the Ph.D. degree from the centre for vision, speech and signal processing, university of Surrey, UK. He is an assistant professor with the Department of Computer Engineering, Bilkent University, Ankara, Turkey and also holds a visiting research fellow position with the centre for vision, speech and signal processing, university of Surrey, UK. His research interests includes secured biometrics, novelty detection and graphical models with applications to image and video analysis.
\end{IEEEbiography}

% if you will not have a photo at all:
\begin{IEEEbiography}[{\includegraphics[scale=.1]{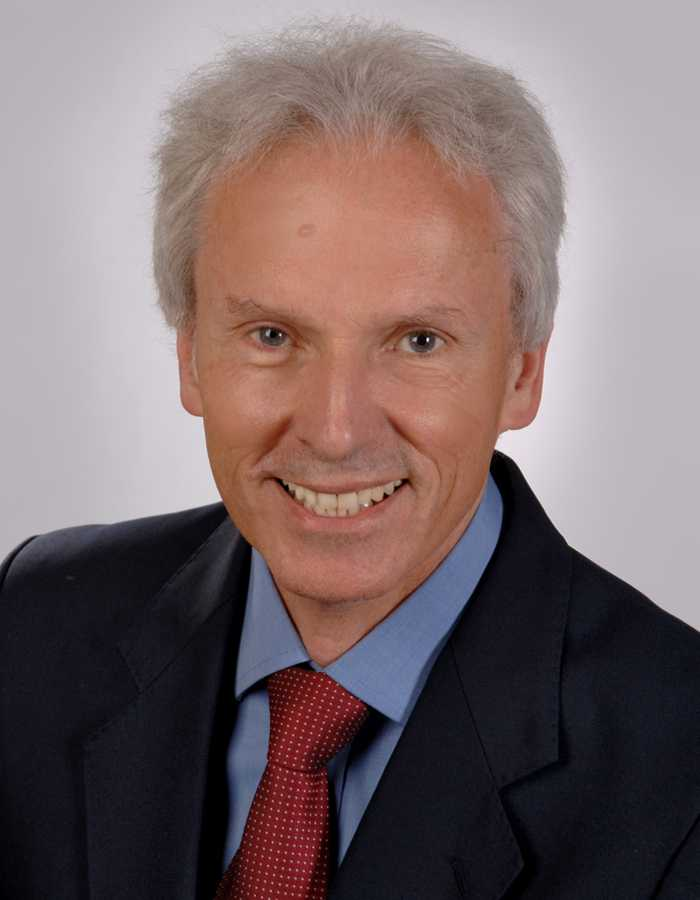}}]{Josef Kittler} (M'74-LM'12) received the B.A., Ph.D., and D.Sc. degrees from the University of Cambridge, in 1971, 1974, and 1991, respectively. He is Professor of Machine Intelligence at the Centre for Vision, Speech and Signal Processing, Department of Electronic Engineering, University of Surrey, Guildford, U.K. He conducts research in biometrics, video and image database retrieval, medical image analysis, and cognitive vision. He published the textbook Pattern Recognition: A Statistical Approach (Englewood Cliffs, NJ, USA: Prentice-Hall, 1982) and over 600 scientific papers. He serves on the Editorial Board of several scientific journals in pattern recognition and computer vision.
\end{IEEEbiography}

% that's all folks
\end{document}